 \documentclass[preprint,3p,12pt]{elsarticle}

\usepackage{graphicx}
\usepackage{amssymb}

\usepackage{amsmath}
\usepackage{booktabs}
\usepackage{xcolor}
\usepackage{caption}
\usepackage{subcaption}
\usepackage{makecell}
\usepackage{float}
\usepackage{siunitx}
\usepackage[labelformat=simple]{subcaption}

\usepackage{enumitem}

\renewcommand{\vec}[1]{\mathbf{#1}}

\journal{Reliability Engineering and System Safety}

\begin{document}
\begin{frontmatter}


\title{Remaining Useful Lifetime Prediction via Deep Domain Adaptation}



\author[paulo]{Paulo Roberto de Oliveira da Costa}
\author[alp]{Alp Ak\c{c}ay}
\author[ying]{Yingqian Zhang}
\author[uzay]{Uzay Kaymak}

\address{Eindhoven University of Technology}
\address{Department of Industrial Engineering}
\address{Eindhoven, The Netherlands}

\begin{abstract}
In Prognostics and Health Management (PHM) sufficient prior observed degradation data is usually critical for Remaining Useful Lifetime (RUL) prediction. Most previous data-driven prediction methods assume that training (source)  and testing (target) condition monitoring data have similar distributions. However, due to different operating conditions, fault modes, noise and equipment updates distribution shift exists across different data domains. This shift reduces the performance of predictive models previously built to specific conditions when no observed run-to-failure data is available for retraining. To address this issue, this paper proposes a new data-driven approach for domain adaptation in prognostics using Long Short-Term Neural Networks (LSTM). We use a time window approach to extract temporal information from time-series data in a source domain with observed RUL values and a target domain containing only sensor information. We propose a Domain Adversarial Neural Network (DANN) approach to learn domain-invariant features that can be used to predict the RUL in the target domain. The experimental results show that the proposed method can provide more reliable RUL predictions under datasets with different operating conditions and fault modes. These results suggest that the proposed method offers a promising approach to performing domain adaptation in practical PHM applications. 
\end{abstract}

\begin{keyword}
Remaining Useful Lifetime \sep Deep Learning \sep Transfer Learning \sep Domain Adaptation


\end{keyword}

\end{frontmatter}


\section{Introduction}
\label{introduction}

Companies interested in increasing the reliability of their assets have been investing in Prognostics and Health Management (PHM) systems to increase reliability, availability and to reduce maintenance costs of engineering assets \cite{atamuradov2017prognostics}. In particular, several works have drawn attention to the task of using the collected data coming from sensors and IoT (Internet of Things) assets to perform prediction of maintenance events, such as fault prognostics, detection and diagnostics \cite{Jardine2006}. In PHM, Remaining Useful Lifetime (RUL) relates to the amount of time left before a piece of equipment is considered not to perform its intended function. Accurate RUL prognostics enable the interested parties to assess an equipment's health status and to plan future maintenance actions, e.g. logistics of personnel, spare parts and services \cite{Papakostas2010}.\par  
For example, in our current industry collaboration, a leading manufacturer of medical imaging systems requires prognostics prediction models that can leverage real-time data collected continuously over many locations. These systems are complex and modelling the precise degradation mechanisms is not possible. Also, although they perform similar procedures, the systems log different multivariate sensor data due to equipment version updates, sensor malfunctioning and timing (equipment installed at a later stage have less temporal data). Furthermore, sensor values have different distributions due to distinct usage and degradation levels. In such cases, high dimensional temporal data has to be directly used to determine the health state of the systems and models have to adapt to incoming changes in the data.\par %
In the PHM literature, physics, statistical and machine learning approaches have been proposed to address the RUL prediction problem. Physics-based approaches build mathematical models that describe the degradation processes of the failure mechanisms \cite{Cubillo2016}. Such models require prior degradation knowledge and provide accurate RUL estimation when failure can be described using its physical properties \cite{Lei2018}. Statistical methods usually attempt to fit the observations under a probabilistic method that can describe the uncertainty of the degradation process \cite{Si2011}. Their shortcomings relate to assumptions about health state transitions and data distributions. On the other hand, machine learning models focus on learning the degradation patterns directly from acquired complex raw data. In general, machine learning models are non-parametric and can be applied in practice even without prior information about the underlying distributions and degradation knowledge \cite{Lei2018}. \par 
Several machine learning methods have been studied for prognostic prediction problems, including methods like Support Vector Machines (SVM) \cite{Dong2013}, Support Vector Regression (SVR) \cite{Benkedjouh2013} and Neural Networks. Neural Networks have been receiving much attention given their ability to approximate high dimensional non-linear functions directly from raw data \cite{Babuska2016}. Moreover, several architectures are specially built to support temporal inputs usually encountered in machine prognostics problems, e.g. Recurrent Neural Networks (RNN). Recently, deep learning methods have been proposed to prognostics problems containing high amounts of time-series input data, including Convolutional Neural Networks (CNN) \cite{Li2018} and variations of RNNs capable of dealing with long input sequences such as the Long Short-Term Memory Networks (LSTM)\cite{ListouEllefsen2019, Zheng2017} and Gated Recurrent Unit (GRU) \cite{Wu2018}.

In classical machine learning, models need enough annotated historical data to be able to train to a significant performance level \cite{Li2018, Zheng2017}. Presumably, interested parties already apply time-based maintenance at their assets and observing run-to-failure behaviours becomes scarcer. To overcome the problem practitioners and researchers have to find ways to handle censored data \cite{hong2009prediction} or generate (e.g. simulate) more data, which leads to imperfect models that do not represent real-world scenarios. Even when enough run-to-failure data are available, algorithms trained for one specific dataset cannot be generalised to a different but related dataset. For example, an algorithm trained for a specific failure mode prediction often does not generalise well to other modes under similar machinery conditions \cite{Lei2018}. Moreover, when the input features change among different equipment versions (e.g. new sensor information is available), it is common practice to retrain the models. This retraining leads to delayed prognostics actions until enough data is available for accurate prediction.\par
To address these issues, predictive models which are trained with specific run-to-failure data, have to adapt to data with different input features, data distributions and limited fault information, i.e. different \textit{domains}. In machine learning, this situation is often referred to as \textit{domain adaptation}. In general, domain adaption methods attempt to solve the learning problem when the main learning goal (\textit{learning task}) is the same, but the domains have different feature spaces or marginal probabilities \cite{Pan2010}. Several algorithms were proposed to different flavours of the domain adaptation problem, which include reducing the domain discrepancy between the source and target via instance re-weighting \cite{jiang2007instance}, subspace alignment \cite{fernando2013unsupervised}, and adversarial deep learning \cite{tzeng2017adversarial, ganin2016domain}. Many of these approaches work well for non-sequential data but are less suitable for multivariate time-series data as they do not usually capture the temporal dependencies present in the data. However, this type of data is prevalent in the maintenance context (e.g. in condition monitoring and equipment output data). Therefore, the general domain adaptation methods are hardly applicable to RUL prognostics.

In this work, we propose to use LSTMs \cite{hochreiter1997long} to address the problem of learning temporal dependencies from time-series sensor data that can be transferred across related RUL prediction tasks with different distributions in their features. We learn from a \textit{source domain} with sufficient run-to-failure annotated data and a \textit{target domain} containing only sensor data. We perform adversarial learning similar to \cite{ganin2016domain} and learn a common domain-invariant feature representation that can be used with the classical backpropagation (through time) algorithm \cite{werbos1990backpropagation}. To the best of our knowledge, we are the first to focus on domain adaptation for RUL prediction regression \textit{task} under varying operating conditions and fault modes.
%
%
Furthermore, we use the C-MAPPS NASA Turbofan degradation datasets \cite{Saxena2008} to validate our results. We chose these datasets as they contain four run-to-failure datasets under different failure modes and operating conditions. As a standard, models built, evaluated and deployed for one particular dataset may perform poorly on unseen datasets with different input feature distributions and fault modes. We show the effectiveness of the proposed method against other adapted and non-adapted models in predicting the RUL of aircraft engines. In practice, capital assets go through re-engineering and redesign during their lifetime to prevent obsolescence problems. In such scenarios, the data from a previous design could be used to adapt learned models to a new design with distinct input data. 

The main contributions of this work include a new model that can handle feature distribution shift across domains under different asset operating conditions and fault modes. Unlike classic domain adaption methods, we incorporate heterogeneous time series data coming from multiple sensors in an RUL regression prediction task. Furthermore, the proposed method can be easily updated as new data becomes available. We show in our experiments that the method improves prognostics predictions on unlabelled target data when compared to non-adapted methods. 

The rest of the paper is structured as follows. In the next section, we briefly discuss the state-of-the-art in prognostics prediction and domain adaptation methods. In the subsequent session, we present our model detailing the learning algorithm and how the temporal dependencies of the data are used to create domain-invariant features. In Section 4, we present the learning procedure and detail the choices of model hyperparameters. In section 5, we compare and contrast the performance of proposed methods using our datasets and provide analysis of the results.

\section{Related Work}
\label{related-work}
\subsection{Machine Learning Methods for RUL Prediction}
In the prognostics literature, several artificial intelligence methods have been proposed to predict the RUL of engineering assets. In particular, authors have proposed several methods that attempt to extract the relationships of acquired sequential data and RUL prediction, such as linear regression \cite{He2008}, Support Vector Regression (SVR) \cite{Benkedjouh2013}, fuzzy-logic systems \cite{zio2010data} and neural networks \cite{tian2012artificial}. 

Neural networks have drawn much attention given its ability to approximate complex functions directly from raw data. For example, \citet{huang2007residual} proposed a Feed Forward Neural Networks (FFNN) architecture in a PHM prediction problems yielding superior results in comparison to other reliability-based approaches. Moreover, in many PHM applications, sequential time-series data are present. Neural Networks architectures such as Recurrent Neural Networks (RNN) are a natural fit for such problems given that their recurrent internal structure can handle sequential patterns in the input data. However, as demonstrated by \citet{bengio1994learning} RNNs have issues when learning long-term dependencies, because of vanishing gradient issues as training progresses. To address these issues Long Short Term Memory (LSTM) \cite{hochreiter1997long} and Gated Recurrent Unit (GRU)\cite{cho2014learning} networks were introduced. Such networks posses internal gates that control how information will flow in the network during the learning procedure. These gates enable the network to preserve its memory state over time and fight the vanishing gradient problem while retaining information for a longer period.

In PHM, \citet{yuan2016} recently showed that LSTMs could outperform RNNs, GRUs and Adaboost-LSTM in an RUL prediction task. \citet{Zheng2017} showed that a sequence of LSTM layers followed by FFNNs could outperform other methods including CNN's in three distinct degradation datasets. \citet{wu2018remaining} presented similar results by extracting features based on a dynamic difference procedure and later training an LSTM for RUL predictions. Results showed that the LSTM also outperforms simpler RNNs and GRU architectures under similar machinery conditions. More recently, \citet{ListouEllefsen2019} showed that Restricted Boltzmann Machines could be used to extract useful weight information by pretraining on degradation data in an unsupervised manner. In this two-stage method, weights extracted in the first step are then used in a further step to fine-tune a supervised LSTM and FFNN model. A genetic algorithm (GA) is used to select the best performing hyperparameters. The methodology holds the state-of-the-art prediction results for the C-MAPSS datasets, presenting it as an effective method for temporal degradation data prediction. 

CNN's are notable for being able to extract spatial information from 2D and 3D high dimensional data yielding the best-known results in several related tasks such as image segmentation, classification and captioning \cite{Hossain2018ACS}. CNN's can also handle 1D sequential data and extract high-level features by combining convolution and max-pooling operations while sliding a local receptive field over the input features. Several CNN architectures have been proposed for remaining useful lifetime prognostics. \citet{babu2016deep} proposed a 2D deep CNN to predict the RUL of a system based on normalised variate time series from sensor signals; they show the effectiveness of the CNN in comparison to Multi-layer Perceptron (MLP), SVR and Relevance Vector Regression (RVR). \citet{Li2018} proposed to apply 1D convolutions in sequence without pooling operations. The results show that the proposed architecture can extract deep features from wear data by concatenating only convolution operations. They show competitive results on the C-MAPPS dataset without incurring in high training times encountered when training recurrent models.

\subsection{Domain Adaptation Methods for PHM}

Most previous studies have focused on predicting the RUL when enough \textit{run-to-failure} data is available and assuming the training and future data come from the same distribution and feature space \cite{Li2018, ListouEllefsen2019, Zheng2017,babu2016deep,  yuan2016}. However, in real-life PHM scenarios, RUL values may be absent and coming from different marginal distributions between training and testing data. Examples include collecting data coming from different devices in varied operating conditions. Moreover, the data can also have different features across training and future data and run-to-failure data can be expensive to obtain. Unsupervised domain adaptation addresses those issues by building algorithms that can be applied when there is \textit{domain shift} in the distribution and feature spaces \cite{Pan2010}. Early methods to unsupervised adaptation attempted to re-weight source example losses to reflect the ones in the target distribution \cite{jiang2007instance,huang2007correcting}. Re-weighting based methods often assume a restricted form of domain shift and selection bias which restricts their applications. Subspace alignment methods \cite{fernando2013unsupervised} attempt to find a linear map that minimises the Frobenius norm between a number of top eingenvectors. However, such methods fail to align the distributions among the two data sources. Maximum Mean Discrepancy (MMD) based methods (e.g., Transfer Component Analysis (TCA)  \cite{pan2011domain}) can be interpreted as moment matching methods and can express arbitrary statistics of the data using kernel tricks. Similarly, CORrelation ALignment (CORAL) \cite{sun2016return} attempts to align the second order statistics between source and target domains. 

Recently, domain-adapted neural networks have been proposed for unsupervised domain adaptation. In general, these methods have attempted to restrain the target error by the source error plus a discrepancy metric between the source and the target domains \cite{Ben-David2010}. For example, \citet{long2015learning, tzeng2014deep} propose to incorporate a the MMD metrics to reduce the discrepancy between domains in classification problems. Similarly, \citet{sun2016deep} propose to incorporate a CORAL loss function for the same purpose. Another approach, based on the theory by \citet{Ben-David2010}, is to use a classification loss (Proxy $A$-distance) to directly confuse between domains \cite{Ben-David2010, ganin2016domain, ajakan2014domain}. Adversarial \cite{goodfellow2014generative} methods have also been proposed to the adaptation task. For example, \citet{tzeng2017adversarial} proposes to pre-train a classifier in the source domain task and use its weights in a an adversarial learning task. In their implementation a target network attempts to confuse a discriminator by generating a representation similar to the source domain. This representation is then used to do inference using the source weights. 

In regression, \citet{cortes2014domain} showed that the discrepancy between domains is a distance for the squared loss when the hypothesis set is the reproducing kernel Hilbert space induced by an universal kernel. \citet{lopez2012semi} proposed a method to factorise a multivariate density into a product of bi-variate copula functions to identify independent changes between domains (i.e., covariate \textit{shift}). Therefore, changes in each of the input features can be detected and corrected to adapt a density model across different learning domains. More recently, \citet{nikzad2018domain} has proposed a method a domain-invariant Partial-Least-Squares Regression using a domain regulariser to align source and target distributions in a latent space. Albeit the current attention in the recent literature, few works have attempted to perform domain adaptation when the input data are composed of time-series data. Similar to our work, \citet{Purushotham2016} proposed a method for time-series domain adaptation using Variational Recurrent Autoencoders and show promising results in a classification task using healthcare data.

In PHM,  \citet{Lu2017} proposed a method for domain adaptation where Maximum-Mean Discrepancy (MMD) discrepancy metric is used to find a common latent space in a classification task. \citet{zhang2017new} proposed to use hand-crafted and raw input features to construct a CNN and wide first-layer kernels model capable of performing domain adaptation in the presence of noisy data. Another CNN approach was proposed by \citet{li2019multi} where an MMD parameterised by multiple kernels is proposed to address the domain shift problem. \citet{xie2016} used feature extraction from time and frequency domains using Transfer Component Analysis (TCA) \cite{pan2011domain} for gearbox fault diagnosis. The results showed that the proposed method could find cross-domain features using data under various operating conditions and yields better results in comparison to other dimensionality reduction methods. Recently, \citet{li2018cross} proposed a deep generative model to generate fault target data using a labelled source domain and an unlabelled target domain. The method proposed an MMD metric and a classification loss in the generation phase to induce a common shared space between the domains. Results showed that the method could be used to solve the original target classification problem in the presence of unlabelled data.

Our work builds upon previous successful works using LSTMs to extract temporal features from time-series sensor data for RUL prediction \cite{yuan2016, Zheng2017, wu2018remaining, ListouEllefsen2019}. To handle the distribution shift and different features among tasks we perform \textit{unsupervised} domain adaptation from a labelled source containing observed RUL vales to an unlabelled target data. This scenario is typical in PHM as usually one is interested in predicting the RUL of an asset before failures are observed. We propose a method that can use the labelled failure data from the source domain to predict the RUL of the target domain under different failure modes and operating conditions. Similar to \cite{ganin2016domain} we use a gradient reversal layer to perform adversarial learning during training and induce a domain-invariant representation. \par
Unlike previous domain adaptation works in PHM that focused on classification tasks, our method focuses on a regression task, where the goal is to determine the number of remaining cycles for aircraft engines. Similar to \cite{ganin2016domain} our method utilises a single neural network capable of learning the source regression task and performing adversarial training. As a standard, adversarial learning is achieved by training two or more networks pitted against each other with contrasting objectives in a two-pass optimisation procedure \cite{goodfellow2014generative}. Therefore, this modification allows for an easier implementation requiring only one architecture. Moreover, weight updates can be performed using the classic backpropagation through time algorithm. Our method uses a classification loss \cite{ajakan2014domain} to induce domain confusion as it has recently been shown it can replace the classical MMD metric in classification instances.
\section{LSTM Deep Adversarial Neural Network}
In this section, we present our domain adaptation model to predict the RUL of assets across domains with different fault modes and operating conditions. We first introduce the problem and the notations used in the paper and then further discuss the proposed method and its components. 

\subsection{Problem Definition}

We denote a source domain $D_\mathcal{S} = \{(\vec{x}^i_\mathcal{S},\vec{y}^i_\mathcal{S})\}_{i=1}^{N_\mathcal{S}}$, containing $N_\mathcal{S}$ training examples, where $\vec{x}^i_\mathcal{S}$ belongs to a feature space $\mathcal{X}_\mathcal{S}$ and denotes a multivariate time-series sensor data of length $T_i$ and $q_\mathcal{S}$ features, i.e.  $\{\vec{x}^i_\mathcal{S} = (x^i_t)_{t=1}^{T_i}\} \in  \mathbb{R}^{q_\mathcal{S} \times T_i} $. Moreover, $\vec{y}^i_\mathcal{S} \in \mathcal{Y_\mathcal{S}}$ denotes the remaining useful lifetime values of length $T_i$ with $\{\vec{y}^i_\mathcal{S} = (y^i_t)\} \in \mathbb{R}^{T_i}_{\geq0}$. Where for each $t \in \{1, 2, ..., T_i\}$, $x^i_t \in \mathbb{R}^{q_\mathcal{S}}$ and $y^i_t \in \mathbb{R}_{\geq0}$ represent the $t$-th measurement of all variables and labels, respectively. Similarly, we assume a target domain $D_\mathcal{T} = \{\vec{x}^i_\mathcal{T}\}_{i=1}^{N_\mathcal{T}}$, where $\vec{x}^i_\mathcal{T} \in \mathcal{X}_\mathcal{T}$ and $\mathcal{X}_\mathcal{T} \in \mathbb{R}^{q_\mathcal{T} \times T_i}$ but no labels are available. 
We assume $D_\mathcal{S}$ and $D_\mathcal{T}$  are sampled from distinct marginal probability distributions $P(X_\mathcal{S}) \neq P(X_\mathcal{T})$. 
Our goal is to learn a function $g$ such that we can approximate the corresponding RUL in the target domain examples at testing time directly from degradation data, i.e. $\vec{y}^i_\mathcal{T} \approx g(\vec{x}^i_\mathcal{T})$. Clearly, our assumption is that the true mapping between input output pairs is somewhat similar across domains for adaptation to be possible. At training time we have access to source training samples and their real valued labels $ \{(\vec{x}^i_\mathcal{S},\vec{y}^i_\mathcal{S})\}_{i=1}^{N_\mathcal{S}}$ and we assume access to training samples from the target domain $\{\vec{x}^i_\mathcal{T}\}_{i=1}^{N_\mathcal{T}}$  (\textit{unsupervised domain adaptation}). We assign a domain label $d_{i} \in \{0, 1\}$  to each $i$-th training example to indicate the domain it originates from and to assist domain classification.

\subsection{Time Windows Processing }
To adapt for different sequence lengths and allow information from past multivariate temporal sequences influence the RUL prediction at a point in time we apply a time window approach for feature extraction. The sequential input is assumed to be $\vec{x}^i = (x^i_t)_{t=1}^{T_i} $ where $T_i$ denotes the size of each sequence length. We define a function $h$ that divides each sequence of size $T_i$ in sequential time windows of size $T_w$, i.e. $h_t(\vec{x}^i) =  \{(x^i_{t-T_w}, ..., x^i_{t-1} )\}_{t= T_w + 1}^{T_i} $. After the transformation, at time $t$ all previous sensor data within the time window $ (x^i_{t-T_w}, ..., x^i_{t-1})$ are collected to form a high-dimensional input vector used to predict $y^i_t$. If $T_i \leq T_w$ we apply zero-padding on the left side of $\vec{x}^i$ until $T_i$ has size $T_w + 1$, this ensures that after the transformation each original time series will have $n_i = {T_i-T_w}$ training samples. Also, we define as $\widetilde{N}_\mathcal{S}$ and $\widetilde{N}_\mathcal{T}$ the updated number of examples after the transformation, that is $\widetilde{N}_\mathcal{S} = \sum_{i=1}^{N_\mathcal{S}} n_i$ and $\widetilde{N}_\mathcal{T} = \sum_{i=1}^{N_\mathcal{T}} n_i$. We maintain $T_w$ fixed across source and target data to allow consistency on the number time steps seen by the network before a prediction, although this parameter could be made flexible between domains.

\subsection{Long Short-Term Memory Neural Network}

One choice of learning function to accommodate temporal relationships between inputs and outputs are LSTMs, which have been studied on prognostics and RUL predictions tasks when enough training data is available \cite{yuan2016, ListouEllefsen2019}. Such networks offer recurrent connections capable of modelling the temporal dynamics of sensor data in prognostics scenarios. Moreover, they control how information flows within the LSTM cells by updating a series of gates capable of learning long-term relationships in the input data.  
\begin{figure}[H]
    \centering
    \resizebox{1.0\columnwidth}{!}{%
    \includegraphics{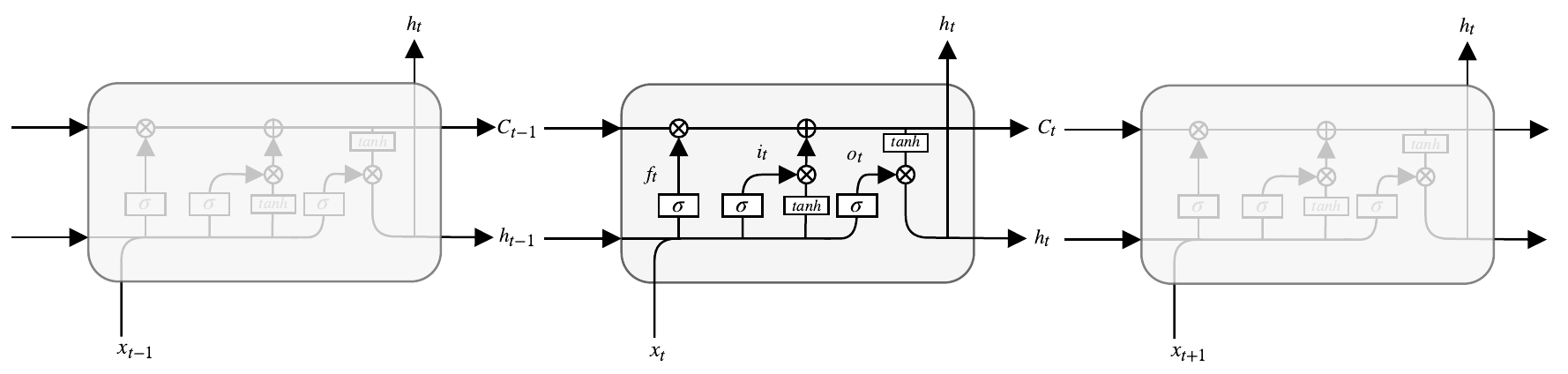}
    }
    \caption{LSTM memory cell \cite{LSTMcell}.}
    \label{fig:lstm+cell}
\end{figure}
In our proposed model, we use LSTM layers to extract temporal features contained in the previous time windows of size $T_w$ before a RUL prediction. In an LSTM, the memory cell (Figure \ref{fig:lstm+cell}) consists of three non-linear gating units that update a cell state $C_t \in \mathbb{R}^h$, using a hidden state vector $h_{t-1} \in \mathbb{R}^h $  and inputs $x^i_t \in \mathbb{R}^q$, where $h$ is the dimension of the LSTM cells and $q$ the input dimension: 
\begin{equation}\label{eq:lstm-f}
    f_t = \sigma\, \big(W_f x^i_t + R_f h_{t-1} + b_f \big)
\end{equation}
\begin{equation}\label{eq:lstm-i}
    i_t = \sigma\, \big(W_i x^i_t + R_i h_{t-1} + b_i \big)
\end{equation}
\begin{equation}\label{eq:lstm-o}
    o_t = \sigma\, \big(W_o x^i_t + R_o h_{t-1} + b_o \big)
\end{equation}
where $\sigma$ is a sigmoid activation function responsible for squeezing the output to the 0-1 range, $W_{g} \in \mathbb{R}^{h \times q} $ are the input  weight matrices,  $R_{g} \in \mathbb{R}^{h \times h}$ are the recurrent weight matrices, and $b_{g} \in \mathbb{R}^h $ are bias vectors. Where the subscript $g$ can either be the forget gate $f$, input gate $i$ or the output gate $o$, depending on the activation being calculated. 

After computing $f_t$, $i_t$ and $o_t$ $\in \mathbb{R}^h$, the new cell state $\widetilde{C}_t$ candidate is computed as follows:
\begin{equation}\label{eq:lstm-ct1}
    \widetilde{C}_t = tanh\, \big(W_C x^i_t + R_C h_{t-1} + b_C \big)
\end{equation}
where, similar to the gate operations: $W_{C} \in \mathbb{R}^{h \times q} $,  $R_{C} \in \mathbb{R}^{h \times h}$, and $b_{C} \in \mathbb{R}^h $.

The previous cell state $C_{t-1}$ is then updated to the new cell state $C_{t}$:
\begin{equation}\label{eq:lstm-ct2}
    C_t = f_t \otimes {C}_{t-1} +  i_t \otimes \widetilde{C}_t
\end{equation}
where $\otimes$ denotes the element-wise multiplication. 

In other words, in the previous equations, the \textit{forget} gate $f_t$ is responsible for deciding which information will be thrown away from the cell state. Next, the \textit{input} gate $i_t$ decides which states will be updated from a candidate cell state. The input and forget gates are then used to update a new cell state for the next time step.

Lastly,  the \textit{output} gate $o_t$ decides which information the cell will output and new hidden state $h_t$ is computed by applying a $tanh$ function to the current cell state times the output gate results.
\begin{equation}\label{eq:lstm-ht}
    h_t = o_t \otimes tanh(C_t)
\end{equation}

\subsection{LSTM Deep Adversarial Neural Network}

\begin{figure*}[h!]
    \centering
    \resizebox{0.8\linewidth}{!}{%
    \includegraphics{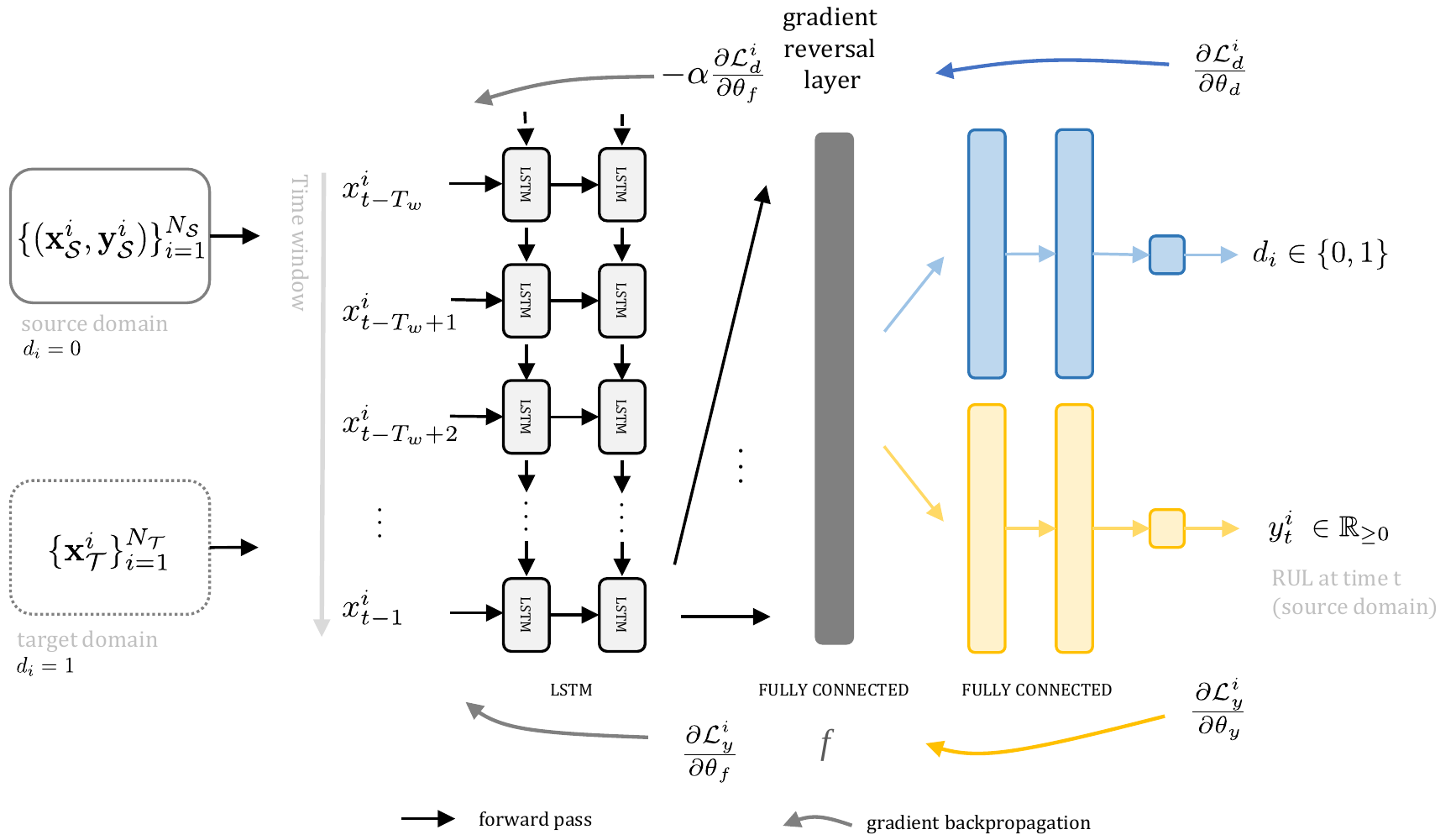}
    }
    \caption{The proposed domain adaptation deep network architecture.}
    \label{fig:architecture}
\end{figure*}

Our model, referred as LSTM-DANN and depicted in Figure \ref{fig:architecture}, is trained to predict for each input $\vec{x}$ a real value $y^i_t$ and its domain label $d_i \in \{0,1\}$. Similar to \cite{ganin2016domain} we use a Domain Adversarial Neural Network (DANN) approach and decompose our learning method in three parts. We assume that the inputs $\vec{x}$ are first decomposed by a combination LSTM layers capable of extracting temporal relationships in the input space to the rest of the network. Our feature extractor $g_{f}$ embeds the inputs in a feature space $f \in \mathbb{R}^l$. We denote the vector of parameters in this layer combination as $\theta_f$, i.e. $f = g_f(x;\theta_f)$. This new feature space $f$ is first mapped to a real-valued $y$ variable via a mapping function $g_y(f ;\theta_y)$ composed of fully connected layers with parameters $\theta_y$. Lastly, the same feature vector $f$ is mapped to a domain label $d_i$ by a mapping function $g_d(f ;\theta_d)$ with parameters $\theta_d$.

During training, we aim at minimising a regression loss $\mathcal{L}_y$ using the observed RUL values from the source domain $\vec{y}_S$. Thus, the parameters of the feature extractor and regressor are optimised towards the same goal, i.e. minimising a regression loss function for the source domain task. This is  performed to ensure that the features $f$ are discriminative towards the main learning task and can be used to predict the RUL at each time $t$. We also aim at finding features that are domain invariant, i.e. we want to find a feature space $f$ in which $P(X_S)$ and $P(X_T)$ are similar. To address this contrasting objective, we look at an auxiliary loss $\mathcal{L}_d$ over the domain classifier function $g_d$. We want to estimate the dissimilarity between domains by inducing a high adversarial loss in the domain-invariant features when the domain classifier $g_d(f ;\theta_d)$ has been trained to discriminate between the two domains.

To enforce such behaviour in the network, we train the model in a two-pass adversarial \cite{goodfellow2014generative} procedure. In the first pass, we learn features by minimising the weights of the feature extractor in the direction of the regression loss $\mathcal{L}_y$. In the second pass, we train the algorithm to maximise the same weights in the direction of a domain-classification loss that is being trained to minimise its overall domain classification error $\mathcal{L}_d$.

In other terms, we define the model loss functions in terms of the learning functions $g$ and parameters $\theta$ and we aim at minimising a combined loss function $\mathcal{L}$ expressed as:
\begin{equation}\label{eq:loss_full}
    \begin{split}
    &\mathcal{L} (\theta_f,\theta_y,\theta_d)  =  \frac{1}{\widetilde{N}_\mathcal{S}} \sum_{i=1}^{\widetilde{N}_\mathcal{S}} \mathcal{L}_y^i(\theta_f,\theta_y) \,  - \\  &\alpha \, 
    \Bigg( \frac{1}{\widetilde{N}_\mathcal{S}} \sum_{i=1}^{\widetilde{N}_\mathcal{S}} \mathcal{L}_d^i(\theta_f,\theta_d) +  
    \frac{1}{\widetilde{N}_\mathcal{T}}
    \sum_{i=1}^{\widetilde{N}_\mathcal{T}} \mathcal{L}_d^i(\theta_f,\theta_d) 
    \Bigg)
    \end{split}
\end{equation}
and the losses $\mathcal{L}_y^i$ and $\mathcal{L}_d^i$ are expressed as:
\begin{equation}\label{eq:MSE}
    \mathcal{L}_y^i(\theta_f,\theta_y) =   \left| \hat{y}_t^i - y_t^i  \right| ^p 
\end{equation}
\begin{equation}
    \mathcal{L}_d^i(\theta_d,\theta_y) = -  \bigg[ d_{i}\log(\hat{d}^{\,i}_t) + (1 - d_{i})\log(1 - \hat{d}^{\,i}_t)\bigg]
\end{equation}
where $ \hat{y}_t^i$ is the RUL prediction at time $t$ coming from the source domain, i.e. $ \hat{y}_t^i = g_y(g_f(h_t(\vec{x});\theta_f);\theta_y)$ and $\hat{d}^{\,i}_t$ is the domain prediction from source and target domains i.e. $\hat{d}^{\,i}_t = g_d(g_f(h_t(\vec{x}) ;\theta_f);\theta_d)$. Where $\mathcal{L}_y^i(\theta_f,\theta_y)$ is a regression loss that take the form of the Mean Absolute Error (MAE) when $p = 1$ and the Mean Squared Error (MSE) when $p = 2$. $\mathcal{L}_d^i(\theta_d,\theta_y)$ is the binary cross-entropy loss between domain labels $d^{\,i}$ and $\alpha$ is a positive hyperparameter that weighs the domain classification loss during training. Both losses are commonly used loss functions in regression and classification problems.   

We optimise the function $\mathcal{L}$ by searching for a saddle point solution $\hat{\theta}_f,\hat{\theta}_y,\hat{\theta}_d,$ of the minimax problem below:
\begin{equation}
(\hat{\theta}_f,\hat{\theta}_y) = \arg\min_{\theta_f,\theta_y} \mathcal{L} (\theta_f,\theta_y,\hat{\theta}_d)
\end{equation}
\begin{equation}
\hat{\theta}_d = \arg\max_{\theta_d} \mathcal{L} (\hat{\theta}_f,\hat{\theta}_y,\theta_d)
\end{equation}
and update the learning weights in the network we use gradient updates \cite{ganin2016domain} of the form:

\begin{equation}\label{eq:gradup1}
    \theta_f \leftarrow \theta_f - \lambda \, \Bigg( \frac{\partial \mathcal{L}^i_y}{\partial \theta_f } - \alpha \frac{\partial \mathcal{L}^i_d}{\partial \theta_f }\Bigg)
\end{equation}
\begin{equation}\label{eq:gradup2}
    \theta_y \leftarrow \theta_y - \lambda \, \Bigg( \frac{\partial \mathcal{L}^i_y}{\partial \theta_y } \Bigg)
\end{equation}
\begin{equation}\label{eq:gradup3}
    \theta_d \leftarrow \theta_d - \lambda \, \Bigg(\alpha  \frac{\partial \mathcal{L}^i_d}{\partial \theta_d } \Bigg)
\end{equation}

We use stochastic estimates of the updates in equations (\ref{eq:gradup1}) - (\ref{eq:gradup3}) via Stochastic Gradient Descent (SGD) and its variants. Where the learning rate $\lambda$ represents the learning steps taken by the SGD algorithm as training progresses. To achieve the desired updates, we use a Gradient Reversal Layer (GRL) \cite{ganin2016domain} alongside the gradient updates. This layer does not perform any changes in the weights of the network during the forward pass. On gradient updates, however, it changes the sign of the gradient of the subsequent levels multiplied by a factor $\alpha$. The GRL makes it possible to learn the weights without many transformations of current implementations of the backpropagation algorithm in common deep learning libraries.

\subsection{Dropout Regularisation}

To effectively learn neural network models one has to account for its capabilities on learning complicated patterns seen in raw data, but also its tendency to overfit the training data. Dropout \cite{srivastava2014dropout} is a regularisation method that can prevent overfitting in deep neural network architectures. It provides a simple solution to the overfitting problem by randomly dropping network units and their connections during training to prevent such units from getting highly specialised in the training data. At testing, a single fully connected network is used to approximate the averaging over all the thinned networks used during training. 

This method significantly reduces overfitting, and it has shown considerable results in many prediction tasks \cite{srivastava2014dropout}. In this work, we apply the dropout method independently in the feature weights $\theta_f$, $\theta_d$ and $\theta_y$. In the feature extraction layers, we want to avoid weights to be too specialised in one of the domains without adapting to changing input data. In the remaining layers, we aim to prevent overfitting on both tasks (regression and classification). We search for the best-performing dropout fraction for our model during the hyperparameter tuning phase.

\section{Design of Experiments}
In this section, we describe the experiments using the proposed model to predict the RUL using degradation data coming from different domains. We describe the datasets used in the experiments and the details about the implementation.  

\subsection{C-MAPPS Datasets}

The method is evaluated using the benchmark Commercial Modular Aero-Propulsion System Simulation (C-MAPPS) \cite{Saxena2008} datasets containing turbofan engine degradation data. The C-MAPPS datasets are composed of four distinct datasets that contain information coming from 21 sensors as well as 3 operational settings. Each of the four datasets possesses a number of degradation engines split into training and testing data. Moreover, the datasets have run-to-failure information from multiple engines collected under various operating conditions and fault modes. 

\begin{table}[!h]
\centering{
\resizebox{0.8\columnwidth}{!}{%
\begin{tabular}{@{}lllll@{}}
\midrule
Data & FD001 & FD002 & FD003 & FD004 \\ \midrule
Engines: Training ($N$) & 100 & 260 & 100 & 249 \\
Engines: Testing & 100 & 259 & 100 & 248 \\
Operating Conditions & \multicolumn{1}{l}{1} & \multicolumn{1}{l}{6} & \multicolumn{1}{l}{1} & 6 \\
Fault Modes & \multicolumn{1}{l}{1} & \multicolumn{1}{l}{1} & \multicolumn{1}{l}{2} & 2 \\ 
\midrule
\end{tabular}
 }}
\caption{The C-MAPPS datasets. Each dataset contains a number of training engines (Engines: Training ($N$)) with \textit{run-to-failure} information and a number of testing engines (Engines: Testing) with information terminating before a failure is observed. Operating Conditions: Each dataset can have one or six (based on altitude (0 - 42000 feet), throttle resolver angle (20 - 100) and Mach (0 - 0.84)) operating conditions.  Fault Modes: Each dataset can have and one (HPC degradation) or two (HPC degradation and Fan degradation) fault modes.}
\label{tab:cmapps}
\end{table}
Engines in the datasets are considered to start with various degrees of initial wear but are considered healthy at the start of each record. As the number of cycles increases the engines begin to deteriorate until they can no longer function. At this point in time the engines are considered unhealthy and cannot perform their intended function. Unlike the training datasets, the testing datasets contain temporal data that terminates some time before a system failure.

The original  prediction task is to predict the RUL of the testing units using the training units \cite{Saxena2008}. We expand on this goal and consider the case when one has enough run-to-failure data under a set of fault modes and operating conditions but wants to apply a learned model to a different dataset, i.e. we validate our results on a different set of operating conditions and fault modes. We motivate such setup on cases found in maintenance prediction scenarios. Where run-to-failure data is available for assets under specific running conditions, but unobserved failure prevents the use of previously learned models in a domain with different conditions and fault modes.

The details about the four datasets are given in Table \ref{tab:cmapps}. We refer to the datasets as FD001, FD002, FD003 and FD004. The operating conditions in the datasets vary between one (sea level) in FD001 and FD003, to six, based on different combinations of altitude (0 - 42000 feet), throttle resolver angle (20 - 100) and Mach (0 - 0.84) in FD002 and FD004. Also, fault modes vary between one (HPC degradation) in FD001 and FD002, and two (HPC degradation and Fan degradation) in FD003 and FD004. For our experiments, we consider each one of the datasets as source and target domains and perform domain adaptation on the different source-target pairs. 

\begin{figure*}[h!]

    \centering
    \begin{subfigure}[b]{0.5\textwidth}
        \includegraphics[width=\textwidth]{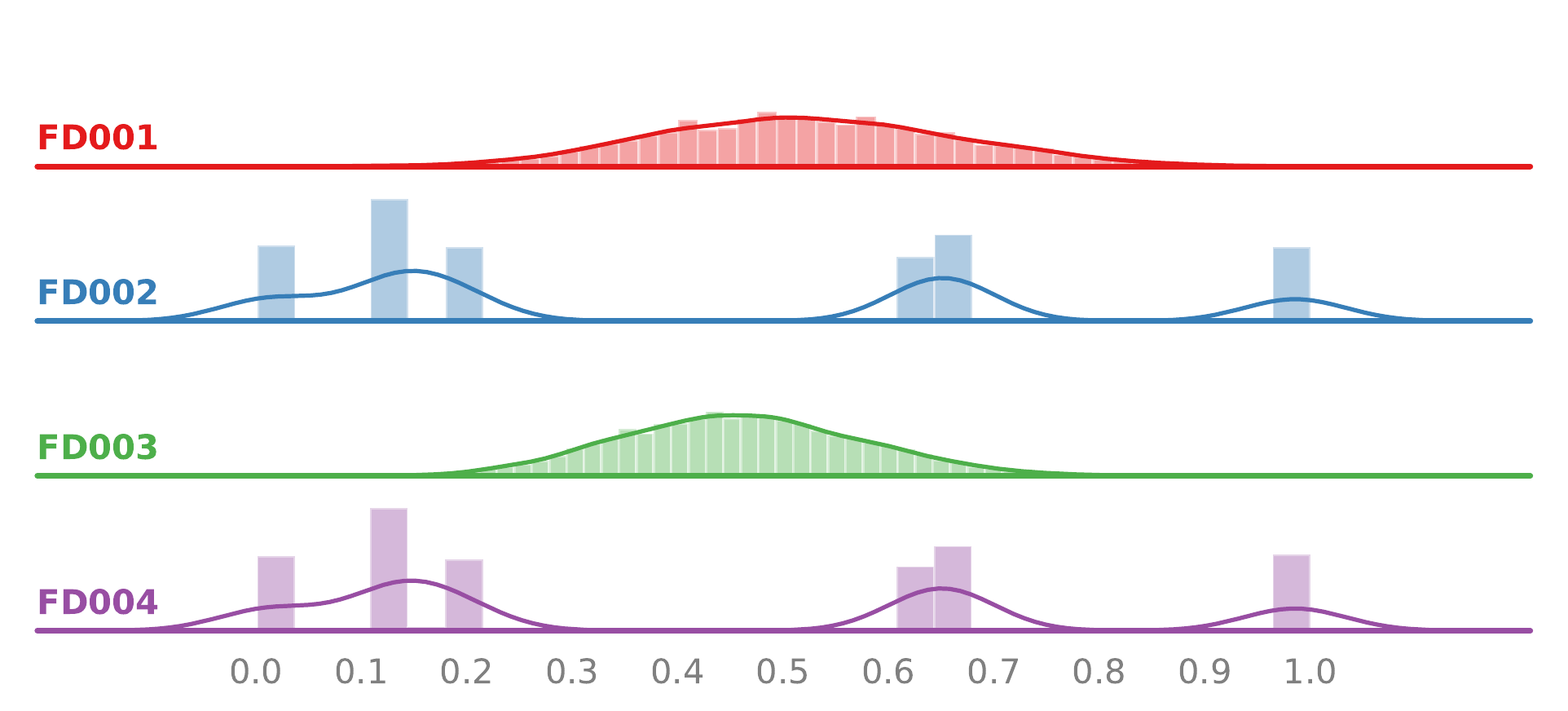}
        \caption{Sensor 2}
         \label{fig:sensor2}
    \end{subfigure}%
    \begin{subfigure}[b]{0.5\textwidth}
        \includegraphics[width=\textwidth]{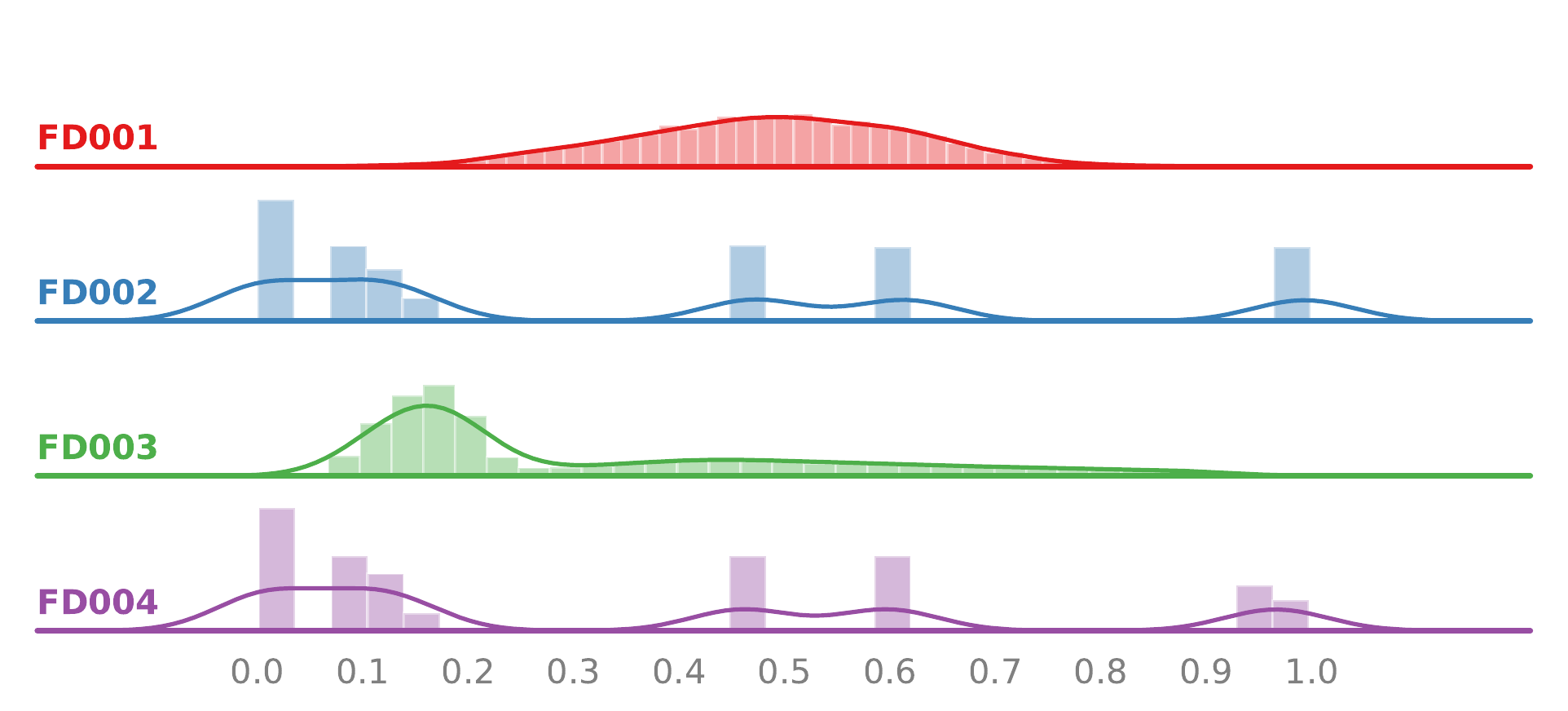}
        \caption{Sensor 7}
         \label{fig:sensor7}
    \end{subfigure}
       \vspace{.5em} 
    \begin{subfigure}[b]{0.5\textwidth}
        \includegraphics[width=\textwidth]{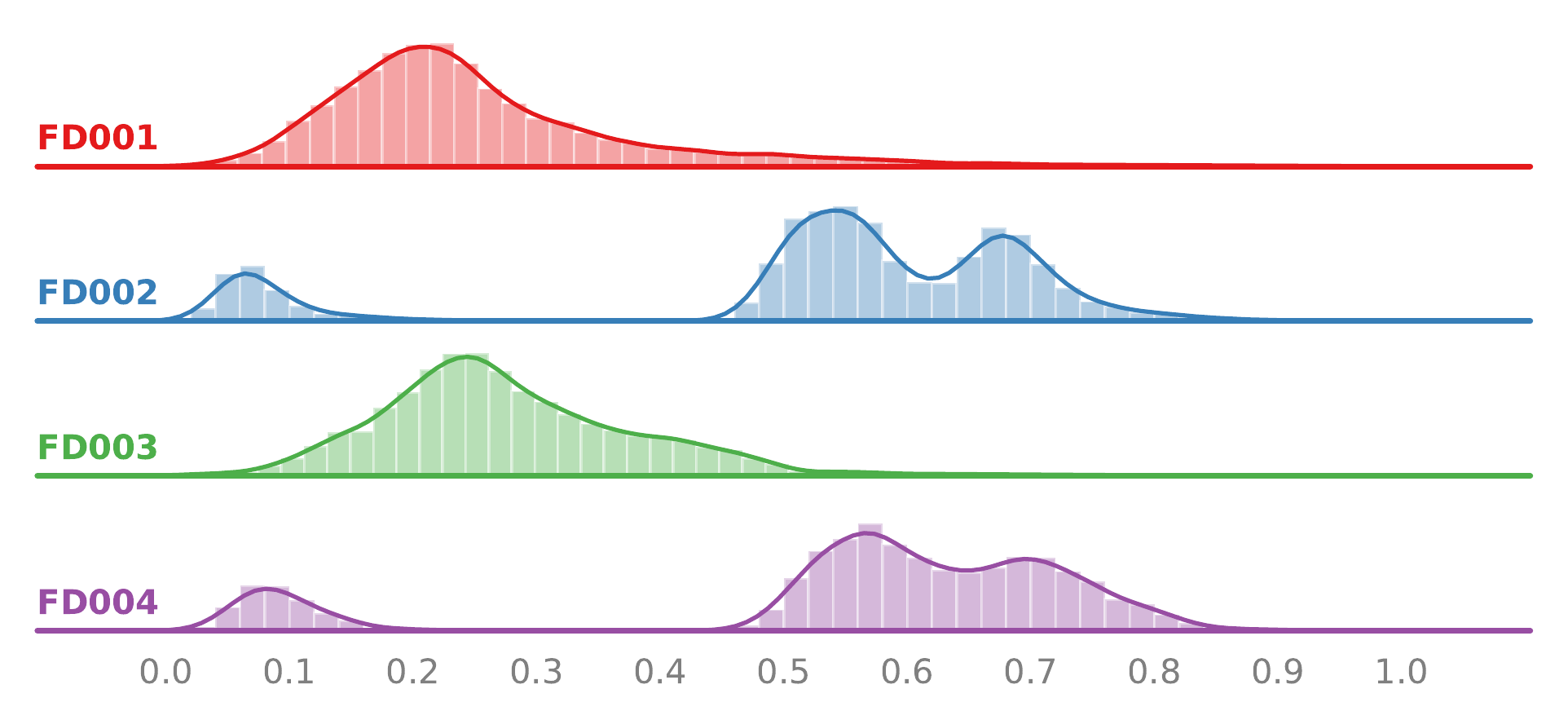}
        \caption{Sensor 14}
         \label{fig:sensor14}
    \end{subfigure}%
    \begin{subfigure}[b]{0.5\textwidth}
        \includegraphics[width=\textwidth]{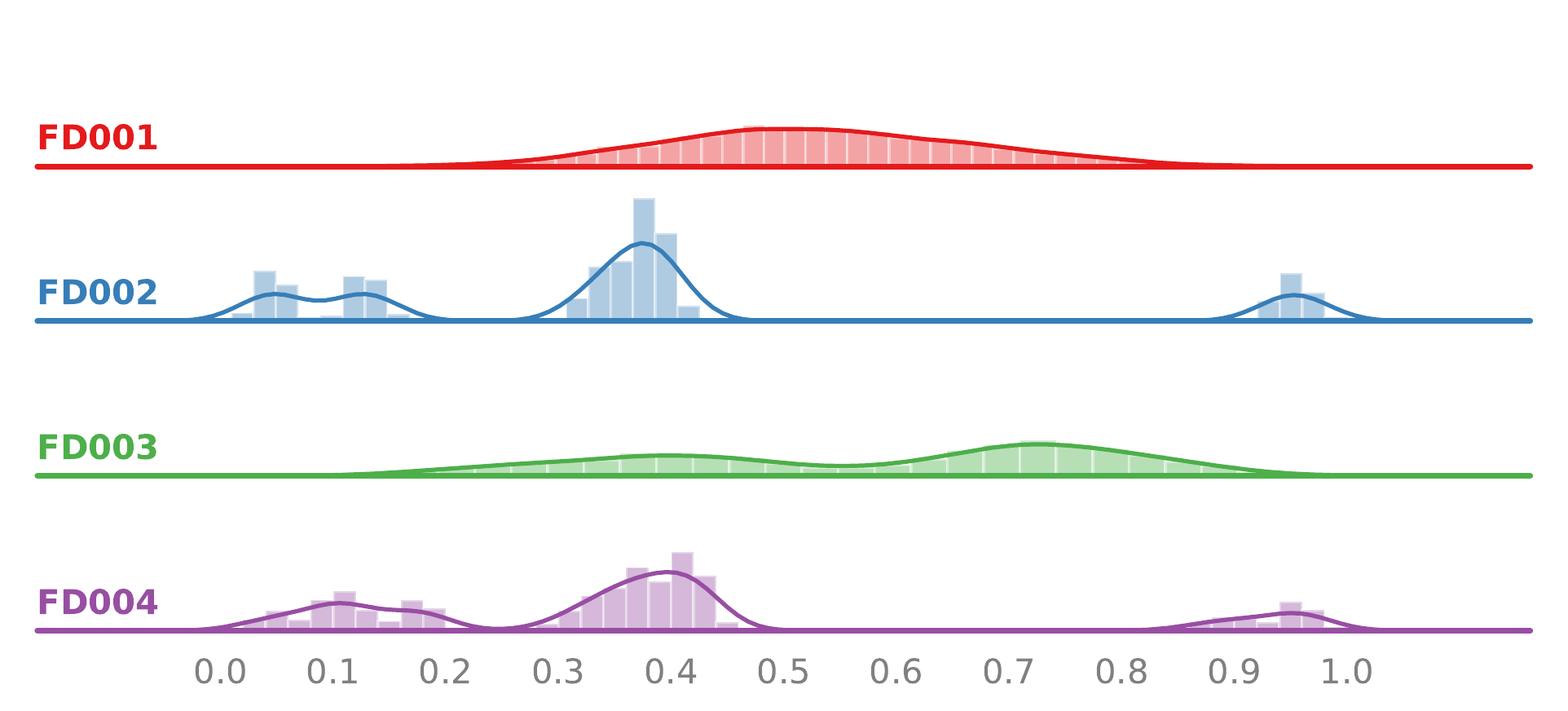}
        \caption{Sensor 15}
         \label{fig:sensor15}
    \end{subfigure}
    \caption{Normalised sensor values 100 time steps before a failure for each C-MAPPS dataset. The sensor distributions are more similar between FD001 and FD003 and FD002 and FD004 pairs due to operating conditions.}\label{fig:sensors}

\end{figure*}

\subsection{Data Preprocessing}

The temporal input data coming from 21 sensor values and 3 operational settings are used across the experiments. We note that for both FD001 and FD003 datasets, 7 sensor values have constant readings. However, as the constant readings are not consistent across the datasets, we keep the sensor values in our experiments to be able to consider their variations in different source and target scenarios. 

Since the original distributions and feature values across the datasets are similar, we need to ensure that enough distribution shift exists so that performing adaptation would make sense. To induce a higher discrepancy between domains and aid gradient descent weight updates, we normalise the input data and RUL values by scaling each feature individually such that it is in the (0-1) range using the min-max normalisation method:
 \begin{equation}
 \label{eq:minmaxnorm}
     norm(x^{i,j}_t)=\frac{x^{i,j}_t-\min(x^j)}{\max(x^j)-\min(x^j)}
 \end{equation}
where $x^{i,j}_t$ denotes the original $i$-th data point of the $j$-th input feature at time $t$ and $x^j$ the vector of all inputs of the $j$-th feature. We perform the normalisation for each dataset individually and perform domain adaptation on the normalised input values.

In RUL prediction tasks it is often not straightforward how to determine the health status and the remaining useful lifetime of an equipment. In our datasets, RUL targets are only available at the last time step for each engine in the test datasets. As \citet{Heimes2008} has shown it is reasonable to estimate the RUL as a constant value when the engines operate in normal conditions. Similar to other works in the literature \cite{ListouEllefsen2019,Lei2018}, we propose to use a piece-wise linear degradation model to define the correct RUL values in the training datasets. That is, after an initial period with constant RUL values, we assume that the RUL targets decrease linearly as the number of observed cycles progresses. We denote as $R_e$ the initial period in which the engines are still working in their desired conditions. A constant $R_e$ of 125 cycles is selected in our experiments to allow comparison to other proposed models in the literature \cite{ListouEllefsen2019,Lei2018}. We point out that the choice of such constant impacts the performance of the RUL prediction methods and that further optimisation can be done to select the best performing $R_e$.

Moreover, we note that the same normalised sensor values coming from distinct datasets can present different distributions according to their degradation level. We show in Figure \ref{fig:sensors} four normalised sensor values of the training examples coming from each of the four datasets 100 time steps before a failure occurs. We observe a lower distribution shift between the dataset pairs FD001, FD003 and FD002, FD004. This is the case because these pairs have data simulated under the same operating conditions, which causes their sensor values to have similar overall distributions near failure \cite{Saxena2008}. However, as it can be seen in Figures \ref{fig:sensor2}, \ref{fig:sensor7} and \ref{fig:sensor15} there are still some distribution shift observed between FD001 and FD003 due to the varying fault modes. Similarly, in Figures \ref{fig:sensor14} and \ref{fig:sensor14} we observe a small distribution shift between FD002 and FD004. In practice, these distribution shifts make models data-specific, i.e. a model trained in one dataset often does not perform well in a different dataset unless the sensor values driving the fault behaviour are similar across source-target pairs.

\subsection{Performance Metrics}

Similar to other prognostic studies using the same datasets, we measure the performance of the proposed method of target datasets using two metrics. We propose to use the Root Mean Squared Error (RMSE) as this can be directly related to equations (\ref{eq:loss_full}) and (\ref{eq:MSE}) and provide an estimation of how well the model is performing in the target prediction task. 
\begin{figure}[h!]
    \centering
    \resizebox{0.6\columnwidth}{!}{%

    \includegraphics{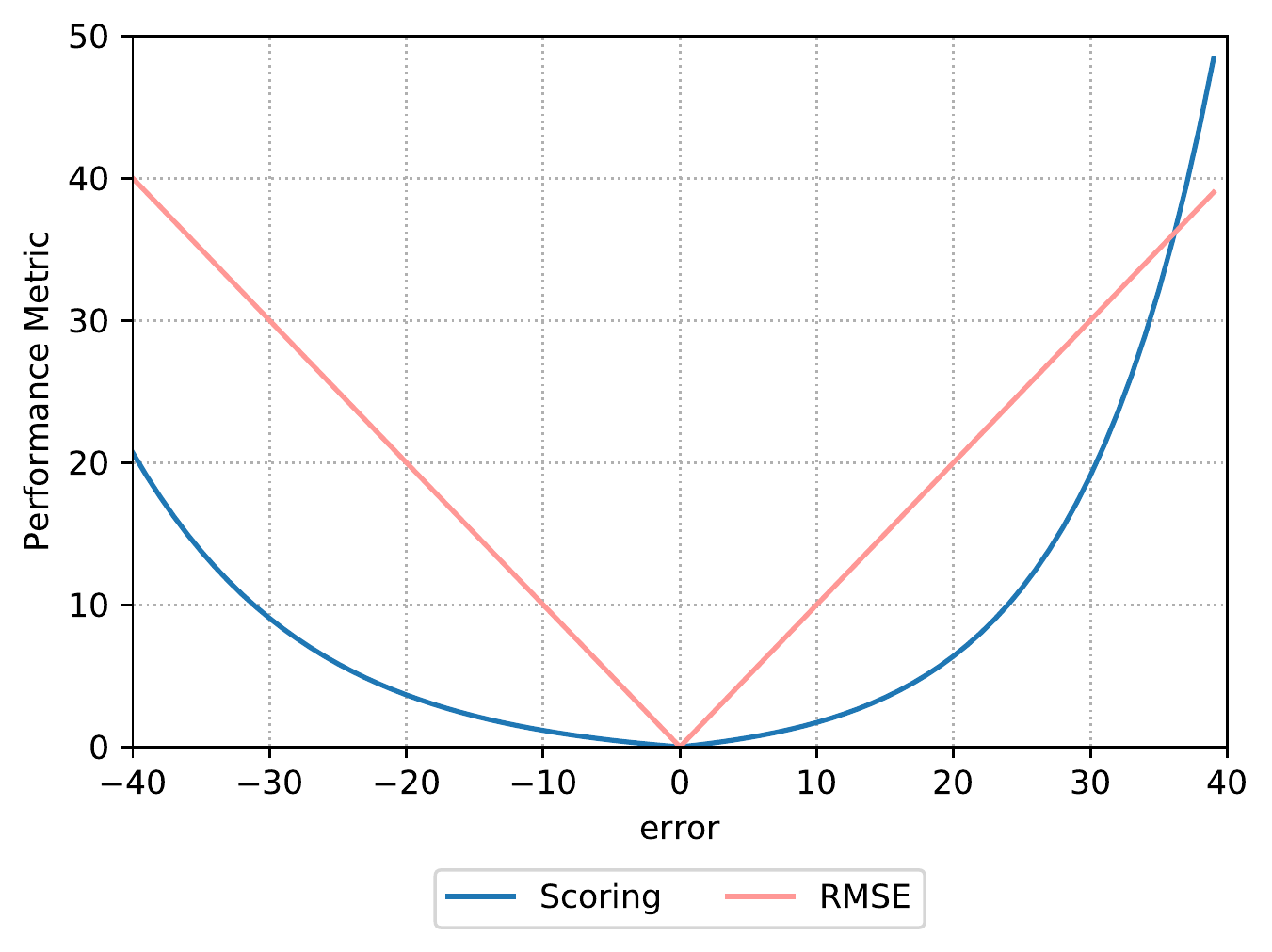}
    }
    \caption{Performance metrics plot. The Scoring performance metric overpenalises positive errors of the RUL prediction.}
    \label{fig:error_f}
\end{figure}

Moreover, we evaluate our model using a scoring function shown in equation (\ref{eq:scoring}) proposed by \citet{Saxena2008}:
\begin{equation}\label{eq:scoring}
  s=\begin{cases}
    \sum_{i=1}^n e^{-\frac{c_i}{a_1}} -1, & \text{if $c_i < 0$}\\
    \sum_{i=1}^n e^{\frac{c_i}{a_2}} -1, & \text{if $c_i \geq 0$}
  \end{cases}
\end{equation}
where $a_1 = 13$ and $a_2 = 10$ and $c_i = \hat{RUL}_i - RUL_i$ \cite{Saxena2008}. That is, $c_i$ is the difference between predicted and observed RUL values. The scoring metric penalises positive errors more than negative errors as these have an impact on RUL prognostics tasks as it can be seen in Figure \ref{fig:error_f}.

\section{Training and Hyperparameter Selection}

\subsection{Training Procedure}

For training, we select one of the four C-MAPPS datasets as source domain and use our proposed domain adaptation method to learn the remaining useful lifetime on the remaining three target datasets. That is, we use the input features and labelled RUL values from the source data, and only the input features values from the target datasets as inputs to the network. The C-MAPPS datasets are normalised individually according to equation (\ref{eq:minmaxnorm}). We apply the time window transformation $h_t$ in both source and target datasets with $T_w = 30$ to allow consistency between the number of time steps a network sees before making a prediction. The selection of the number of time steps is based on previous literature using the same datasets \cite{Lei2018, ListouEllefsen2019}. No further feature engineering is performed in the input data as we aim to extract features automatically using the proposed method. $L_2$ regularisation is applied in the weights $\theta_d$ and $\theta_y$ in equation \ref{eq:loss_full}. Also, we separate the original training data into training (seen by the algorithm) and cross-validation (used for stopping criteria) data containing 90\% and 10\% of the original dataset. 
\begin{figure}[h!]
    \centering
    \includegraphics[width=0.6\columnwidth]{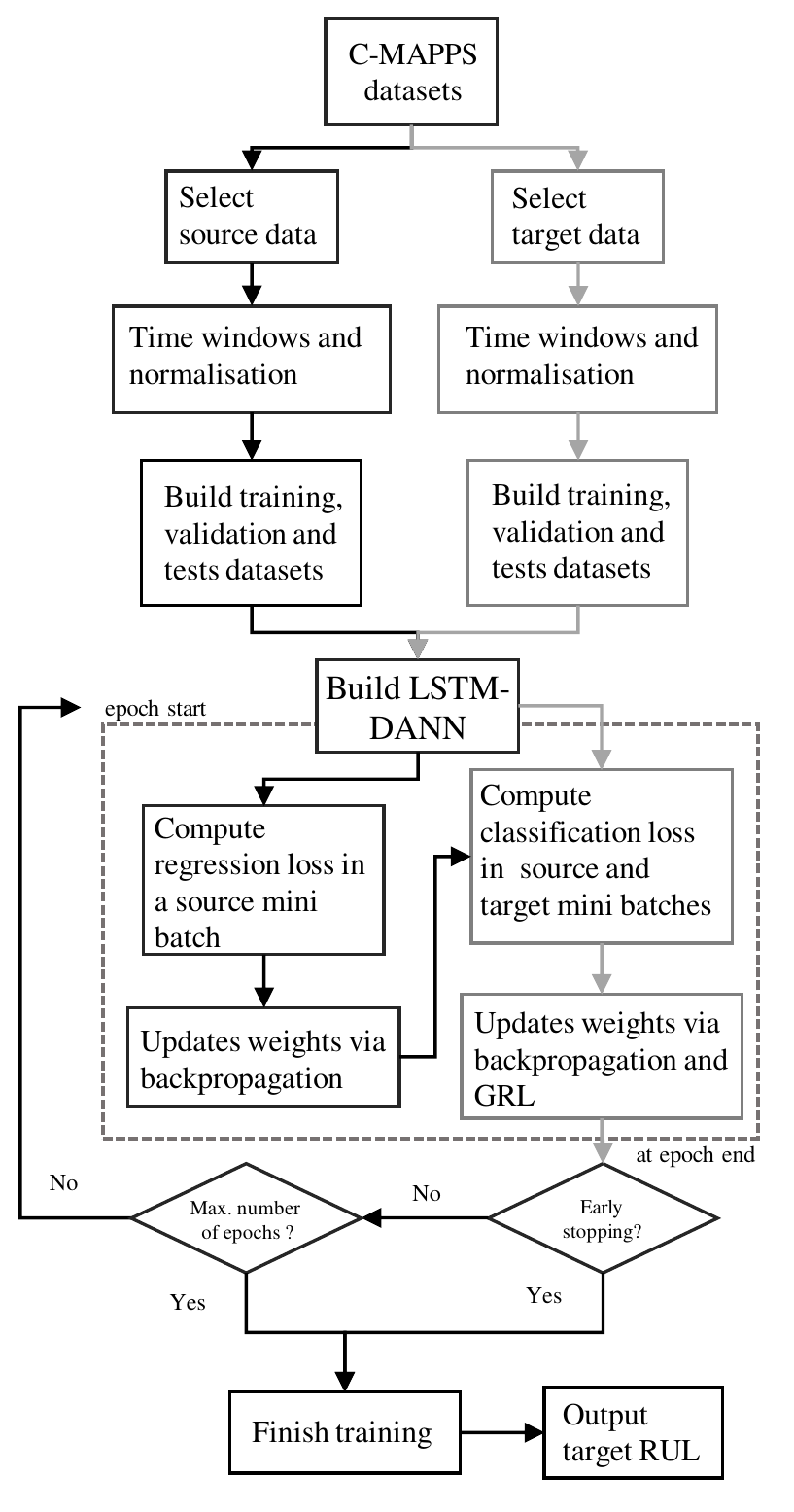}
    \caption{The training procedure of the LSTM-DANN}
    \label{fig:training procedure}
\end{figure}
We split the training dataset into mini-batches (collection of data samples) of data that are used to calculate model error and update model coefficients. During training, we randomly select mini-batches coming from source and target domains to update the weights in the network on each gradient pass. As it can be seen in Table \ref{tab:cmapps} the datasets have different number of training samples. To cope with this difference between domains, we over sample the smaller dataset to match the same number of mini-batches coming from the larger dataset. After the network has seen all training examples coming from both domains we consider an \textit{epoch} finished. Further, we define $d_i = 0$ if the training example comes from the source domain and $d_i = 1$ if it comes from the target domain.

Next, the proposed LSTM-DANN architecture is defined including the number of LSTM and fully connected hidden layers, number of cells in each layer, learning rate and gradient update algorithm. We train using the Rectified Linear Unit (ReLU) as activation function. The SGD and RMSProp \cite{tieleman2012lecture} algorithms are used to update the weights in the network. Our implementation breaks the learning in two models sharing the feature extraction layers. One model aims at learning the regression task using the source domain sensor and RUL information. The other model aims at finding domain-invariant features using the adversarial loss function in equation (\ref{eq:loss_full}). We point out that the second model is the one responsible for the adversarial learning aspect of the model as it includes the GRL to swap the gradient signals. While the weights $\theta_d$  are updated to minimise the classification loss between domains, the weights $\theta_f$ are updated to maximise the domain classifier loss at the same time. The complete diagram of the learning procedure can be seen in Figure \ref{fig:training procedure}.

We train the models for a maximum of 200 epochs and interrupt training if no improvement is seen for 20 epochs. In our case the MAE, $p=1$ in equation (\ref{eq:MSE}), presents the best performing results.  In addition, a varying learning rate is adopted, we start with a fixed learning rate, and after 100 epochs the learning rate is multiplied by a 0.1 factor to allow for stable convergence. We clip the norm values of the gradients to 1 in the SGD algorithm to avoid exploding gradients. Finally, the data coming from the target domain including the RUL values are fed to the network to calculate final RUL estimations and the performance measures can be obtained. 
\subsection{Hyperparameter Selection}
We perform grid search on more sensitive hyperparameters: optimiser (opt) and learning rates (lr), and fine-tune the remaining parameters manually. The range considered for each hyperparameter is shown in Table \ref{grid-search}. To asses the quality of the proposed algorithm we need to validate the hyperparameters without using the RUL values coming from the target domain. In our proposed methodology, we evaluate the performance of the adaptation task by observing the cross-validation error and the domain classifier error on the source domain. In general, we observed that performance results are better when a lower source error is achieved while the domain classification stabilises in loss values that lead to an accuracy close to a random guess. We select the hyperparameters that yield the lowest source RMSE. We report the resulting hyperparameters settings in Table \ref{hyperparam}.\par
\begin{table}[h!]
\centering{
\resizebox{0.8\columnwidth}{!}{%
\begin{tabular}{ll}
\midrule
Hyperparameter                       & Range                       \\ \midrule
Learning rate (source regression)            & \{0.001, 0.01, 0.1\}        \\
Learning rate (domain classification)           & \{0.001, 0.01, 0.1\}        \\
Batch size                          & \{256, 512, 1024\}          \\
Optimiser                        & \{SGD, RMSProp\}            \\
Number of layers (LSTM)                & \{1, 2\}                    \\
Number of units (LSTM)               & \{32, 64, 100, 128\}             \\
Number of units ($f$)                 & \{30, 32, 64, 128, 512\}              \\
Number of layers (source regression)         & \{1, 2\}                    \\
Number of units (source regression)         & \{16, 20, 32, 64, 128\}              \\
Number of layers (domain classification)        & \{1, 2\}                    \\
Number of units (domain classification)        & \{16, 20, 32, 64, 128\}              \\
$L_2$ Regularisation                    & \{0.0, 0.01, 0.1\} 
\\
$\alpha$                             & \{0.8, 1.0, 2.0, 3.0\} \\ \midrule

\end{tabular}%
}}
\caption{Hyperparameter values evaluated in the proposed methodology.}
\label{grid-search}
\end{table}

\begin{table*}[h!]
\centering
\resizebox{\textwidth}{!}{%
    \begin{tabular}{cllllllllllllll}
    \hline
    Source: FD001 & Target & \begin{tabular}{@{}l@{}}$L_2$ \\ Regularisation\end{tabular}  & \begin{tabular}{@{}l@{}}LSTM \\ Layers (units)\end{tabular}  & \begin{tabular}{@{}l@{}}LSTM \\ Dropout\end{tabular} & Units $f$ & \begin{tabular}{@{}l@{}}Source regression  \\ Layers (units)\end{tabular}  & \begin{tabular}{@{}l@{}}Source regression \\ Dropout\end{tabular} &\begin{tabular}{@{}l@{}}Domain classification \\ Layers (units)\end{tabular}  & \begin{tabular}{@{}l@{}}Domain classification \\ Dropout\end{tabular} & $\alpha$ & batch size & lr source reg. & lr domain class. & opt \\
    \hline
    - & FD002 & 0.01 & 1 (128) & 0.5 & 64 & 1 (32) & 0.3 & 1 (32) & 0.3 & 0.8 & 256 & 0.01 & 0.01 & SGD \\
    - & FD003 & 0.01 & 1 (128) & 0.5 & 64 & 1 (32) & 0.3 & 1 (32) & 0.3 & 0.8 & 256 & 0.01 & 0.01 & SGD \\
    - & FD004 & 0.01 & 1 (128) & 0.7 & 64 & 2 (32, 32) & 0.3 & 1 (32) & 0.3 & 1.0 & 256 & 0.01 & 0.1 & SGD \\
    \hline
    Source: FD002 & Target & \begin{tabular}{@{}l@{}}$L_2$ \\ Regularisation\end{tabular}  & \begin{tabular}{@{}l@{}}LSTM \\ Layers (units)\end{tabular}  & \begin{tabular}{@{}l@{}}LSTM \\ Dropout\end{tabular} & Units $f$ & \begin{tabular}{@{}l@{}}Source regression  \\ Layers (units)\end{tabular}  & \begin{tabular}{@{}l@{}}Source regression \\ Dropout\end{tabular} &\begin{tabular}{@{}l@{}}Domain classification \\ Layers (units)\end{tabular}  & \begin{tabular}{@{}l@{}}Domain classification \\ Dropout\end{tabular} & $\alpha$ & batch size & lr source reg. & lr domain class. & opt \\
    \hline
    - & FD001 & 0.01 & 1 (64) & 0.1 & 64 & 1 (32) & 0.0 & 2 (16, 16) & 0.1 & 1.0 & 512 & 0.01 & 0.01 & SGD \\
    - & FD003 & 0.01 & 1 (64) & 0.1 & 512 & 2 (64, 32) & 0.0 & 2 (64, 32) & 0.1 & 2.0 & 256 & 0.1 & 0.1 & SGD \\
    - & FD004 & 0.01 & 2 (32, 32) & 0.1 & 32 & 1 (32) & 0.0 & 1 (16) & 0.1 & 1.0 & 256 & 0.1 & 0.1 & SGD \\
    \hline
    Source: FD003 & Target & \begin{tabular}{@{}l@{}}$L_2$ \\ Regularisation\end{tabular}  & \begin{tabular}{@{}l@{}}LSTM \\ Layers (units)\end{tabular}  & \begin{tabular}{@{}l@{}}LSTM \\ Dropout\end{tabular} & Units $f$ & \begin{tabular}{@{}l@{}}Source regression  \\ Layers (units)\end{tabular}  & \begin{tabular}{@{}l@{}}Source regression \\ Dropout\end{tabular} &\begin{tabular}{@{}l@{}}Domain classification \\ Layers (units)\end{tabular}  & \begin{tabular}{@{}l@{}}Domain classification \\ Dropout\end{tabular} & $\alpha$ & batch size & lr source reg. & lr domain class. & opt \\
    \hline
    - & FD001 & 0.01 & 2 (64, 32) & 0.3 & 128 & 2 (32, 32) & 0.1 & 2 (32, 32) & 0.1 & 2.0 & 256 & 0.01 & 0.01 & SGD \\
    - & FD002 & 0.01 & 2 (64, 32) & 0.3 & 64 & 2 (32, 32) & 0.1 & 2 (32, 32) & 0.1 & 2.0 & 256 & 0.01 & 0.01 & SGD \\
    - & FD004 & 0.01 & 2 (64, 32) & 0.3 & 64 & 2 (32, 32) & 0.1 & 2 (32, 32) & 0.1 & 2.0 & 256 & 0.01 & 0.01 & SGD \\
    \hline
    Source: FD004 & Target & \begin{tabular}{@{}l@{}}$L_2$ \\ Regularisation\end{tabular}  & \begin{tabular}{@{}l@{}}LSTM \\ Layers (units)\end{tabular}  & \begin{tabular}{@{}l@{}}LSTM \\ Dropout\end{tabular} & Units $f$ & \begin{tabular}{@{}l@{}}Source regression  \\ Layers (units)\end{tabular}  & \begin{tabular}{@{}l@{}}Source regression \\ Dropout\end{tabular} &\begin{tabular}{@{}l@{}}Domain classification \\ Layers (units)\end{tabular}  & \begin{tabular}{@{}l@{}}Domain classification \\ Dropout\end{tabular} & $\alpha$ & batch size & lr source reg. & lr domain class. & opt \\
    \hline
    - & FD001 & 0.01 & 1 (100) & 0.5 & 30 & 1 (20) & 0.0 & 1 (20) & 0.1 & 1.0 & 512 & 0.01 & 0.01 & SGD \\
    - & FD002 & 0.01 & 1 (100) & 0.5 & 30 & 1 (20) & 0.0 & 1 (20) & 0.1 & 1.0 & 512 & 0.01 & 0.01 & SGD \\
    - & FD003 & 0.01 & 1 (100) & 0.5 & 30 & 1 (20) & 0.0 & 1 (20) & 0.1 & 1.0 & 512 & 0.01 & 0.01 & SGD \\
    \midrule   
\end{tabular}%
}
\caption{Hyperparameters for each source-target experiment pair.}
\label{hyperparam}
\end{table*}

We run the experiments  presented in this paper in a machine running an Intel Core i5 7th generation processor with 16 GB RAM and a GeForce GTX 1070 Graphics Processing Unit (GPU). We implement the models using the Python 3.6 programming language and the Keras \cite{chollet2015keras} deep learning library with the TensorFlow \cite{tensorflow2015-whitepaper} backend.  

\section{Experimental Results}

In this section, the prognostic performance of the proposed domain adaptation method for RUL estimation is presented. All experiments consider each of one the C-MAPPS datasets as source domain and the remaining datasets as target domains. In total we have 12 different experiments and results are averaged over 10 trials for each experiment to reduce the effect of randomness. For each experiment we report the mean and standard deviations of each model's performance. 

We start by comparing the proposed method with baseline LSTM models trained in the source domain and applied on the target domain (SOURCE-ONLY). Also, we compare to models trained in the target domain using the target domain labels (TARGET-ONLY) representing the ideal situation when target RUL values are available for prediction. Furthermore, we assess our domain adaptation method against other popular methods that attempt to align source and target domains before a prediction model is constructed. We compare against Feed Forward Neural Network (FFNN) models trained on the Transfer Component Alignment (TCA) and CORrelation Alignment (CORAL) domain-invariant spaces. We present the methodologies' effectiveness in finding representations that can both adapt to the different wear distributions across domains.

Lastly, we show that our TARGET-ONLY models can be effectively used to predict the RUL values for each of the C-MAPPS datasets. We compare our methodology with the current state-of-the-art methods to assess the general effectiveness of our proposed method when both sensor and RUL information are available for training.

\subsection{Comparison to Non-adapted Models under Domain Shift}

\begin{figure*}
    \setlength{\belowcaptionskip}{-12pt}
    \centering
    
    Source: FD004
    \begin{subfigure}[b]{0.33\textwidth}
        \includegraphics[width=\textwidth]{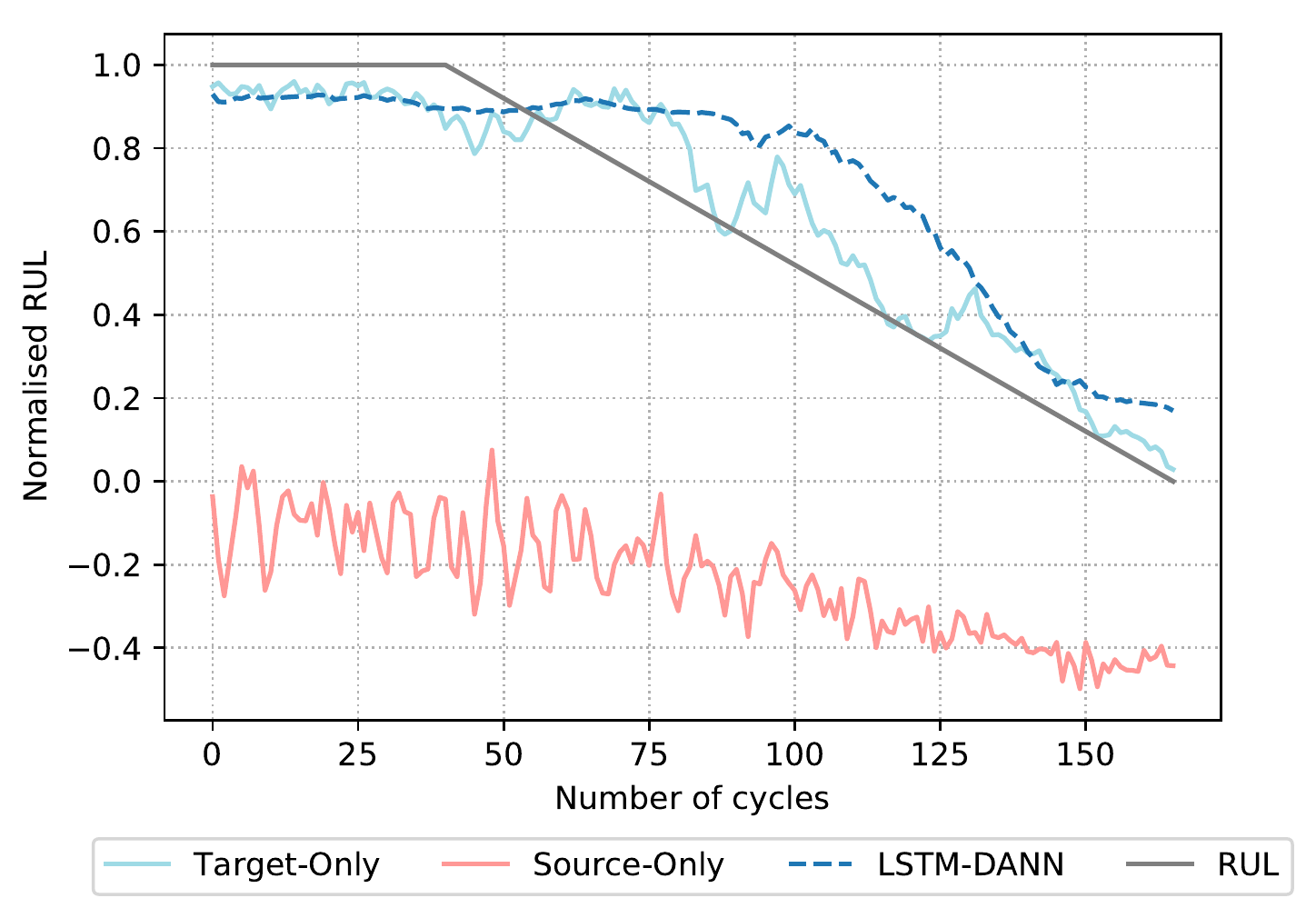}
        \caption{Target: FD001}
         \label{fig:fd04_f01}
    \end{subfigure}%
    \begin{subfigure}[b]{0.33\textwidth}
        \includegraphics[width=\textwidth]{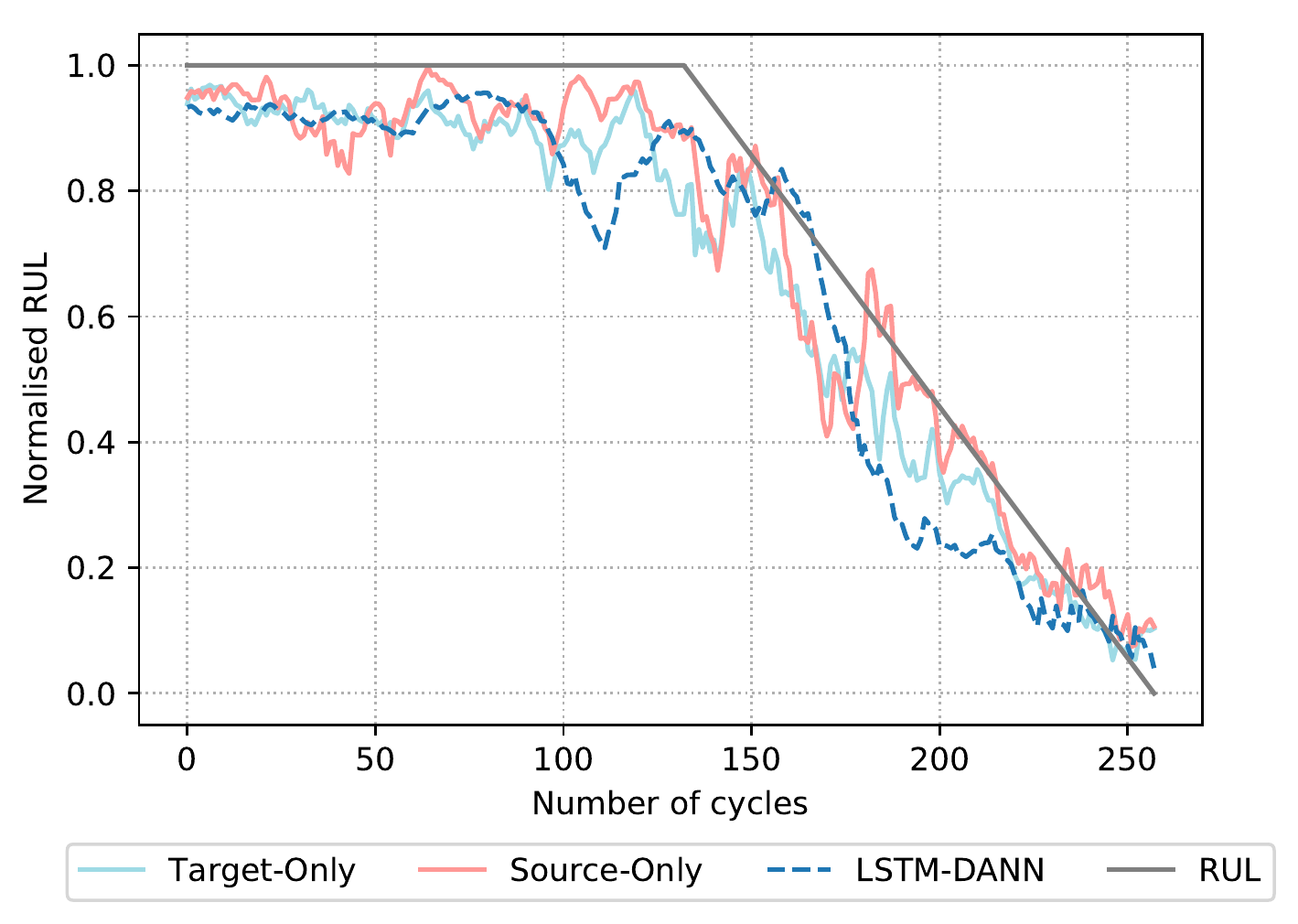}
        \caption{Target: FD002}
         \label{fig:fd04_f02}
    \end{subfigure}%
    \begin{subfigure}[b]{0.33\textwidth}
        \includegraphics[width=\textwidth]{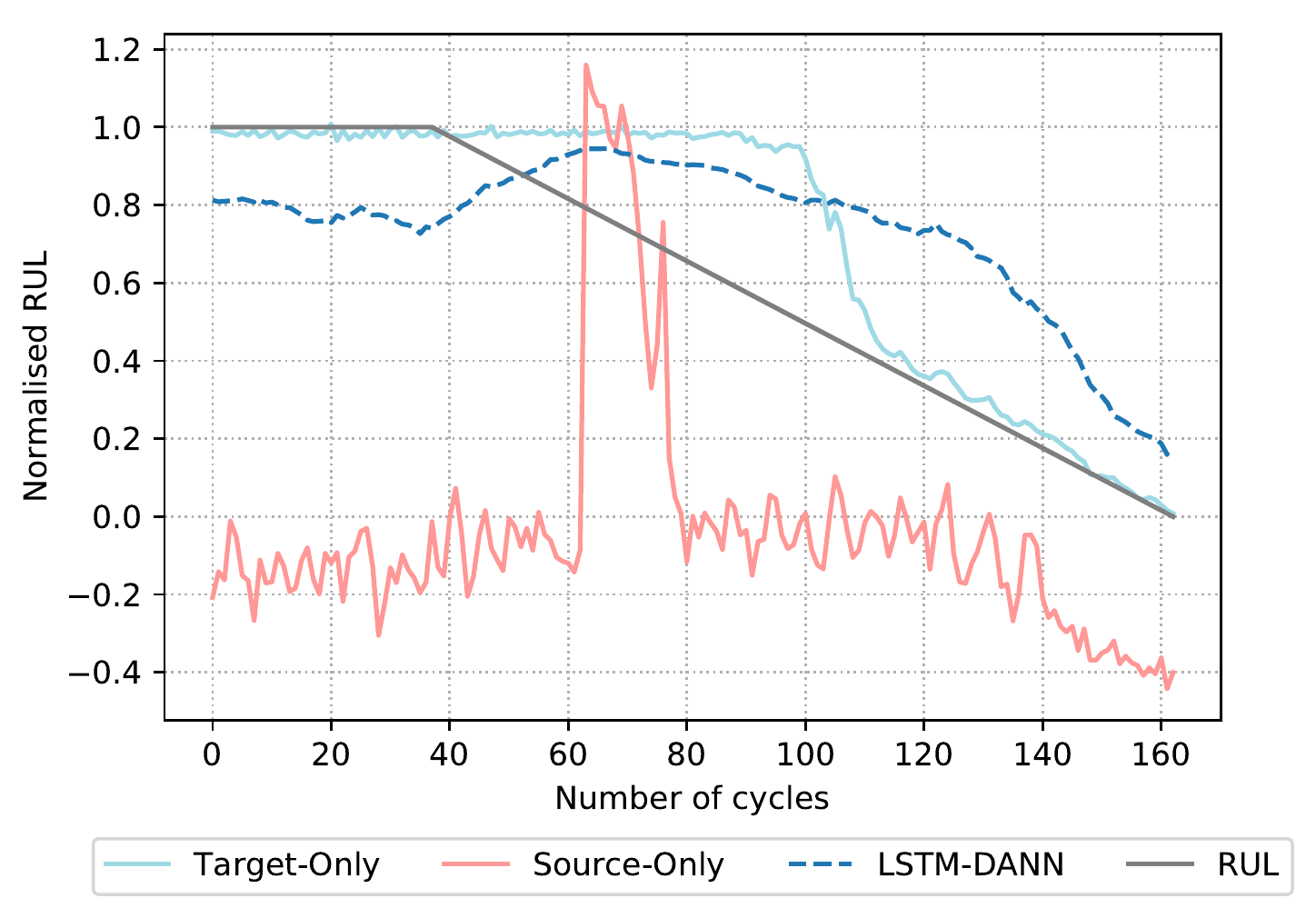}
        \caption{Target: FD003}
         \label{fig:fd04_f03}
    \end{subfigure}
    \vspace{0.1in}\\
    Source: FD003
    \begin{subfigure}[b]{0.33\textwidth}
        \includegraphics[width=\textwidth]{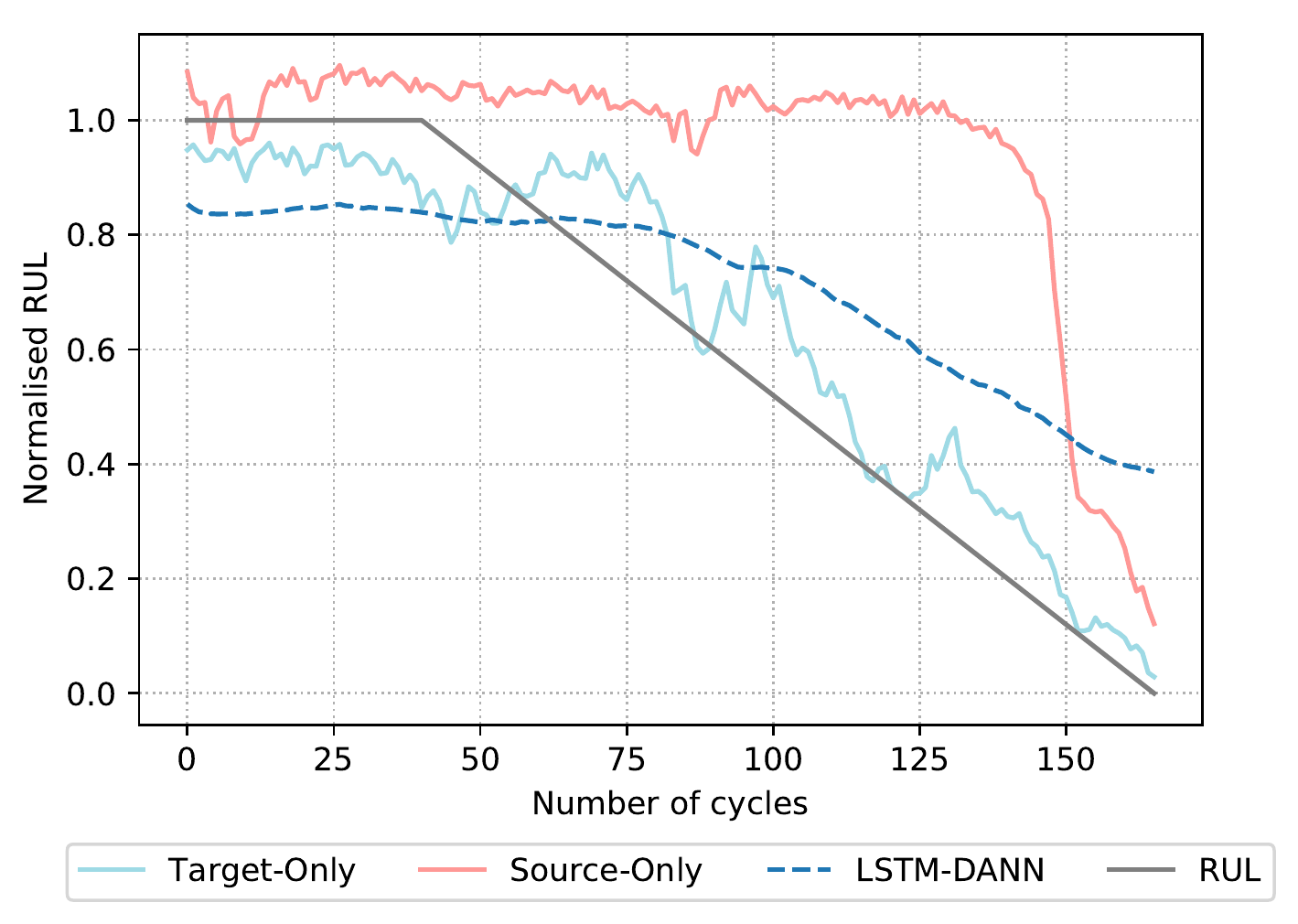}
        \caption{Target: FD001}
         \label{fig:fd03_f01}
    \end{subfigure}%
    \begin{subfigure}[b]{0.33\textwidth}
        \includegraphics[width=\textwidth]{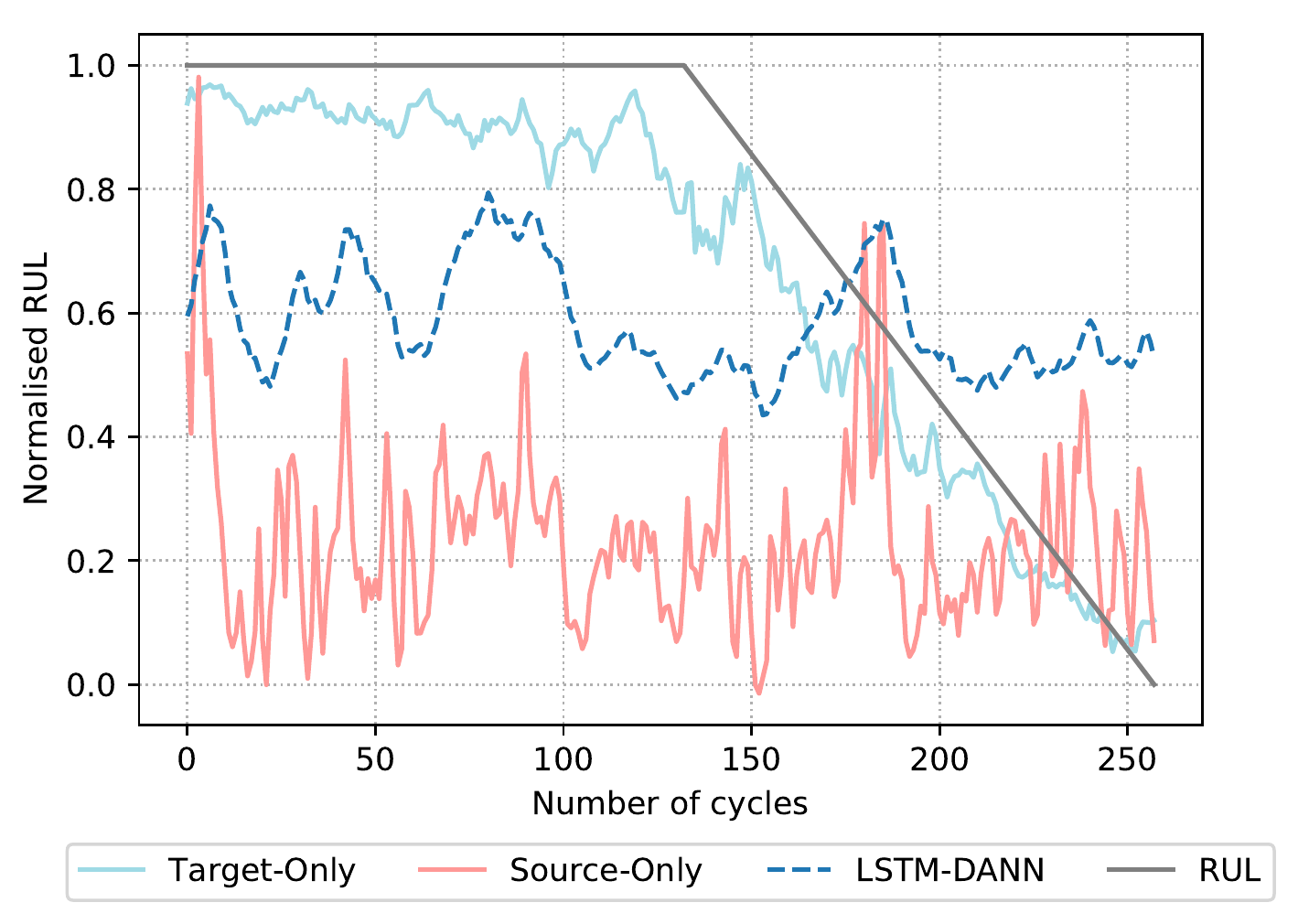}
        \caption{Target: FD002}
         \label{fig:fd03_f02}
    \end{subfigure}%
    \begin{subfigure}[b]{0.33\textwidth}
        \includegraphics[width=\textwidth]{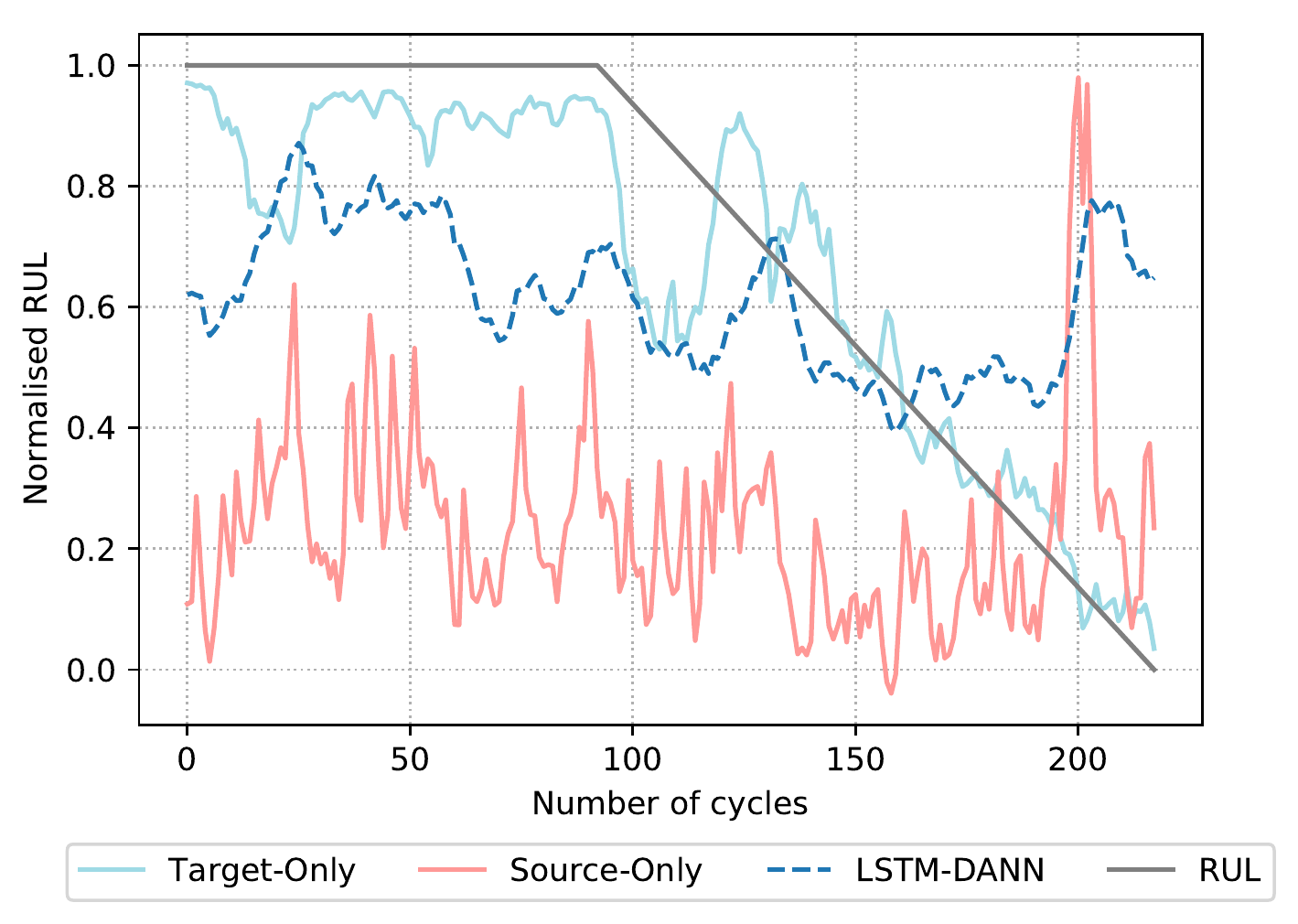}
        \caption{Target: FD004}
         \label{fig:fd03_f04}
    \end{subfigure}
    \hfill
    \vspace{0.1in}\\
    Source: FD002
    \begin{subfigure}[b]{0.33\textwidth}
        \includegraphics[width=\textwidth]{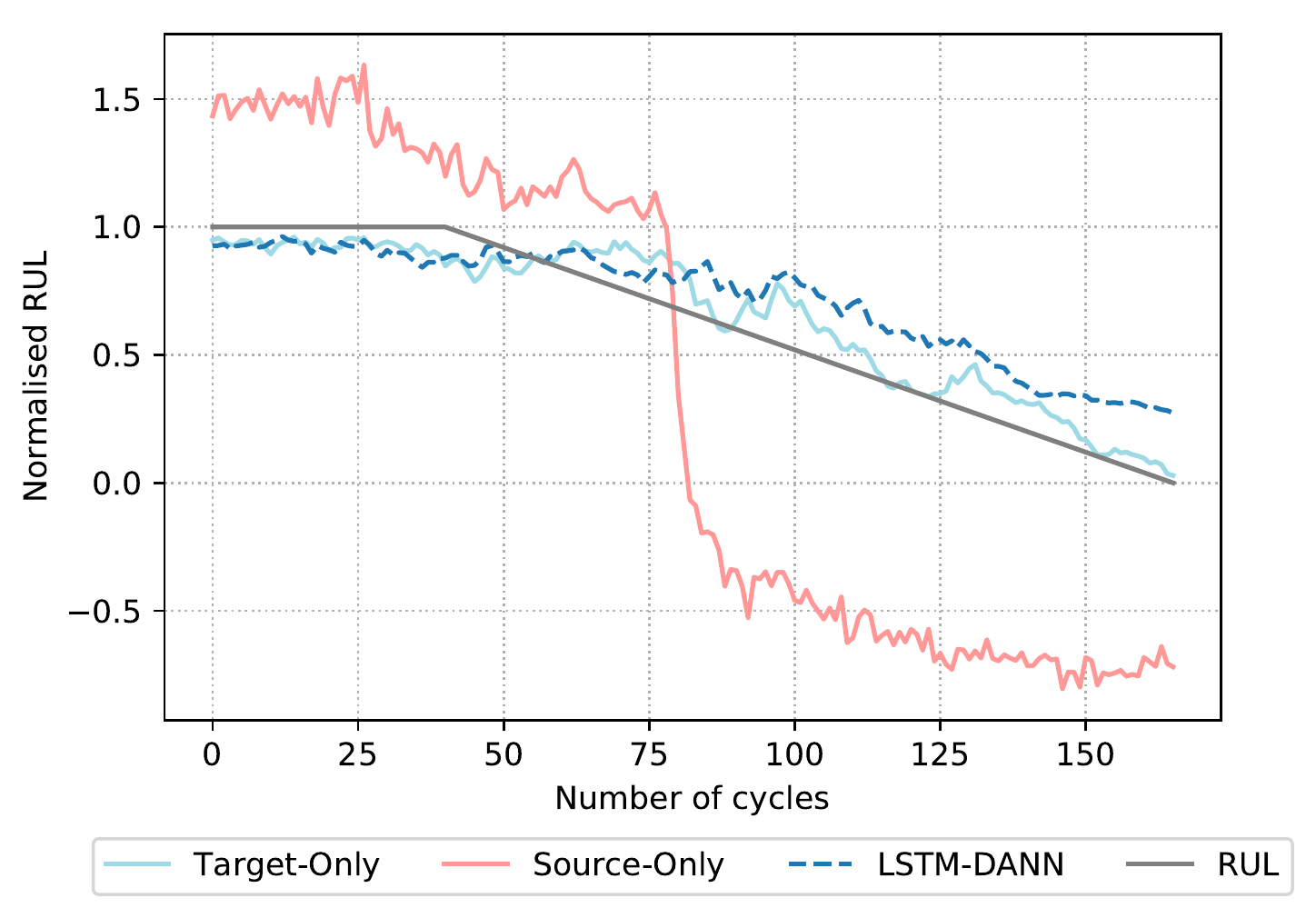}
        \caption{Target: FD001}
         \label{fig:fd02_f01}
    \end{subfigure}%
    \begin{subfigure}[b]{0.33\textwidth}
        \includegraphics[width=\textwidth]{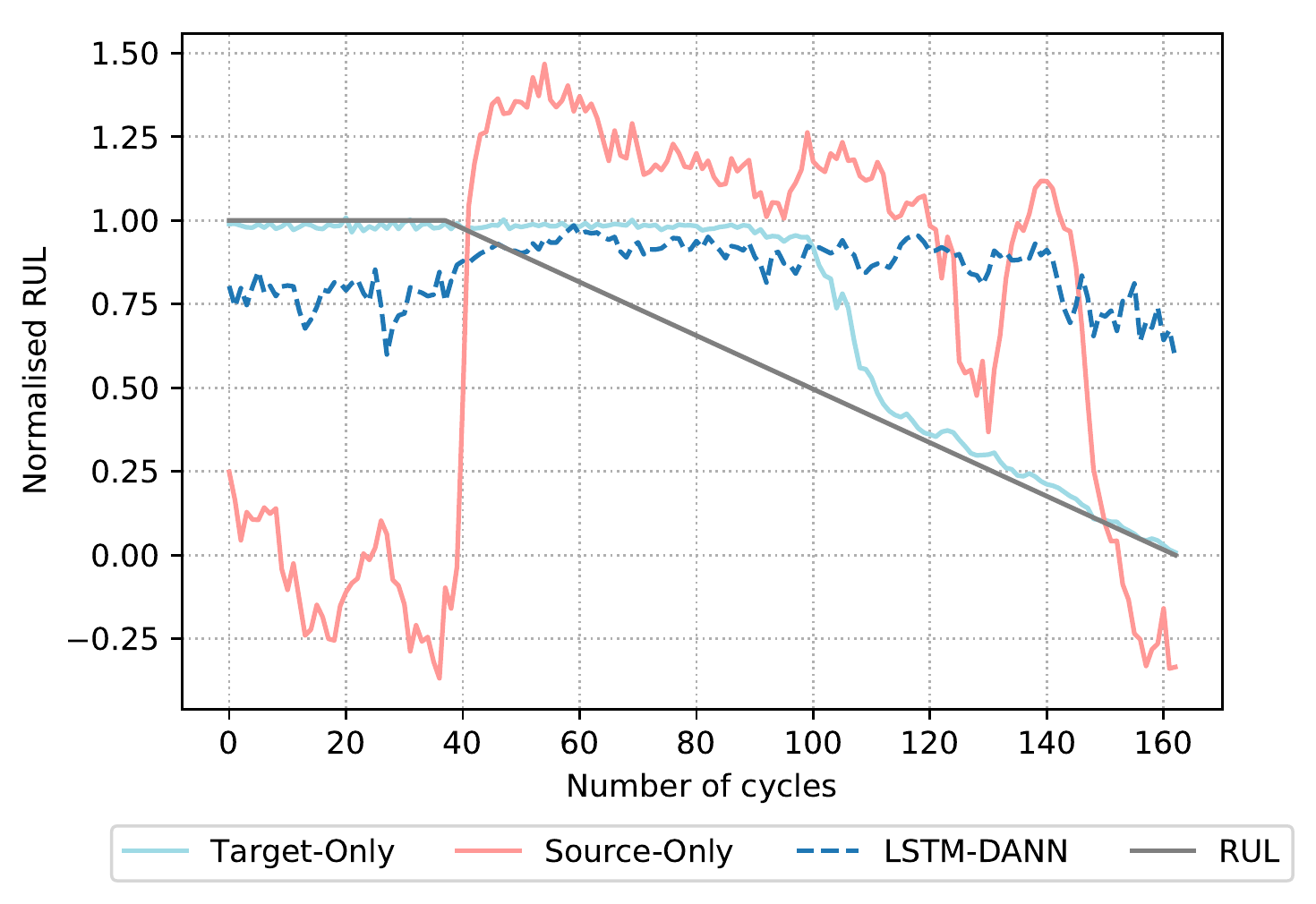}
        \caption{Target: FD003}
         \label{fig:fd02_f03}
    \end{subfigure}%
    \begin{subfigure}[b]{0.33\textwidth}
        \includegraphics[width=\textwidth]{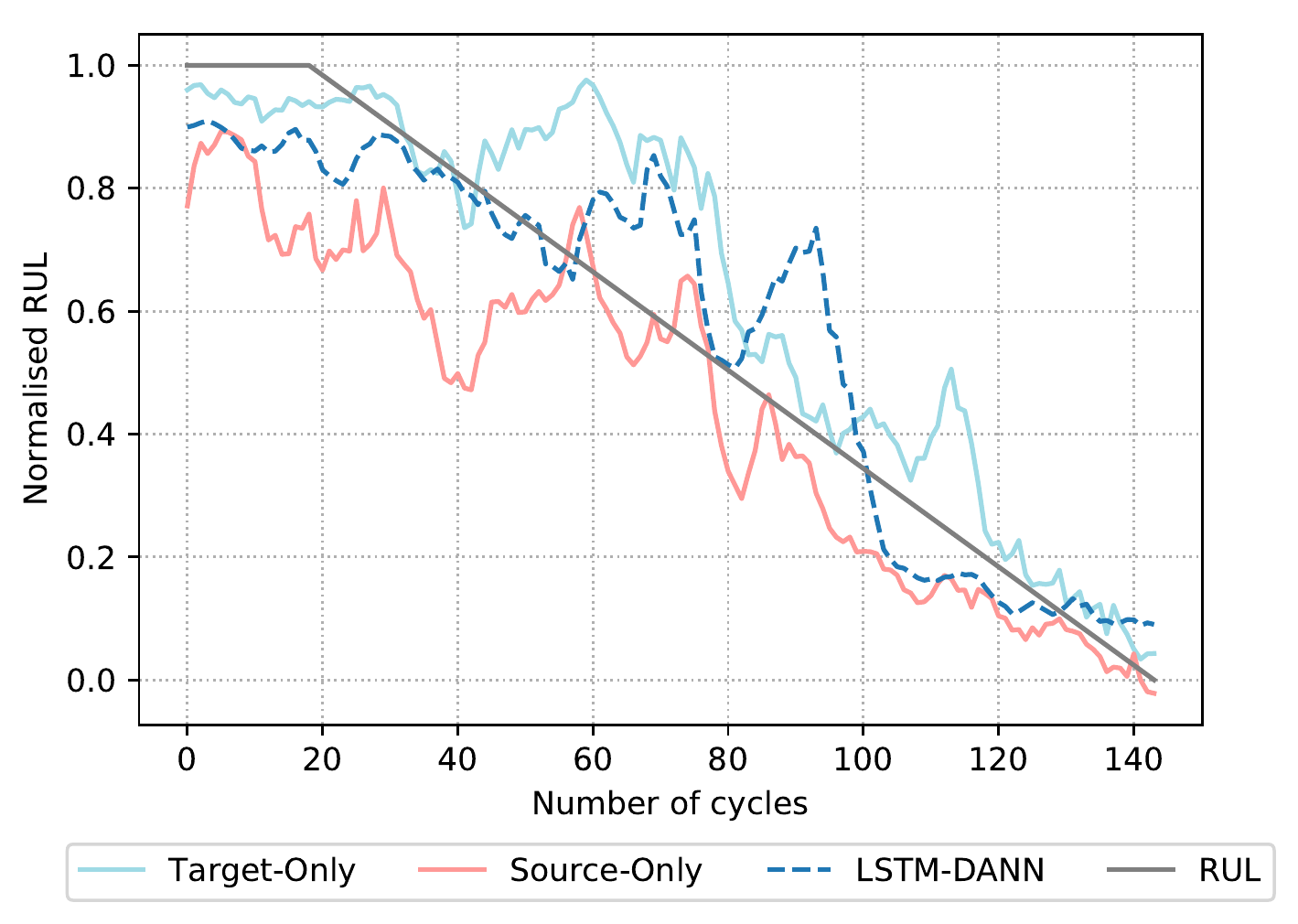}
        \caption{Target: FD004}
         \label{fig:fd02_f04}
    \end{subfigure}
    \hfill
    \vspace{0.1in}\\
     Source: FD001
    \begin{subfigure}[b]{0.33\textwidth}
        \includegraphics[width=0.95\textwidth]{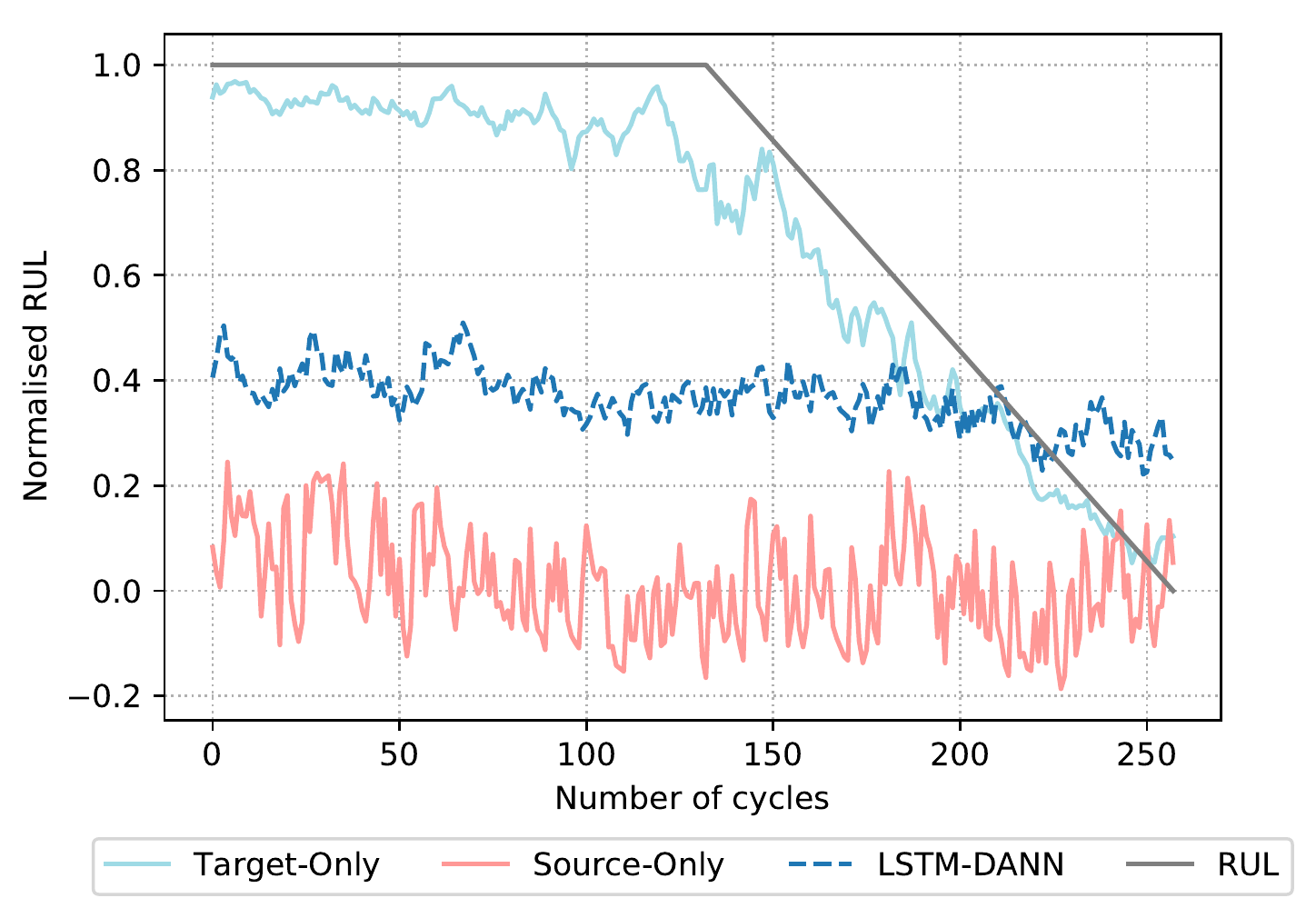}
        \caption{Target: FD002}
         \label{fig:fd01_f02}
    \end{subfigure}%
    \begin{subfigure}[b]{0.33\textwidth}
        \includegraphics[width=0.95\textwidth]{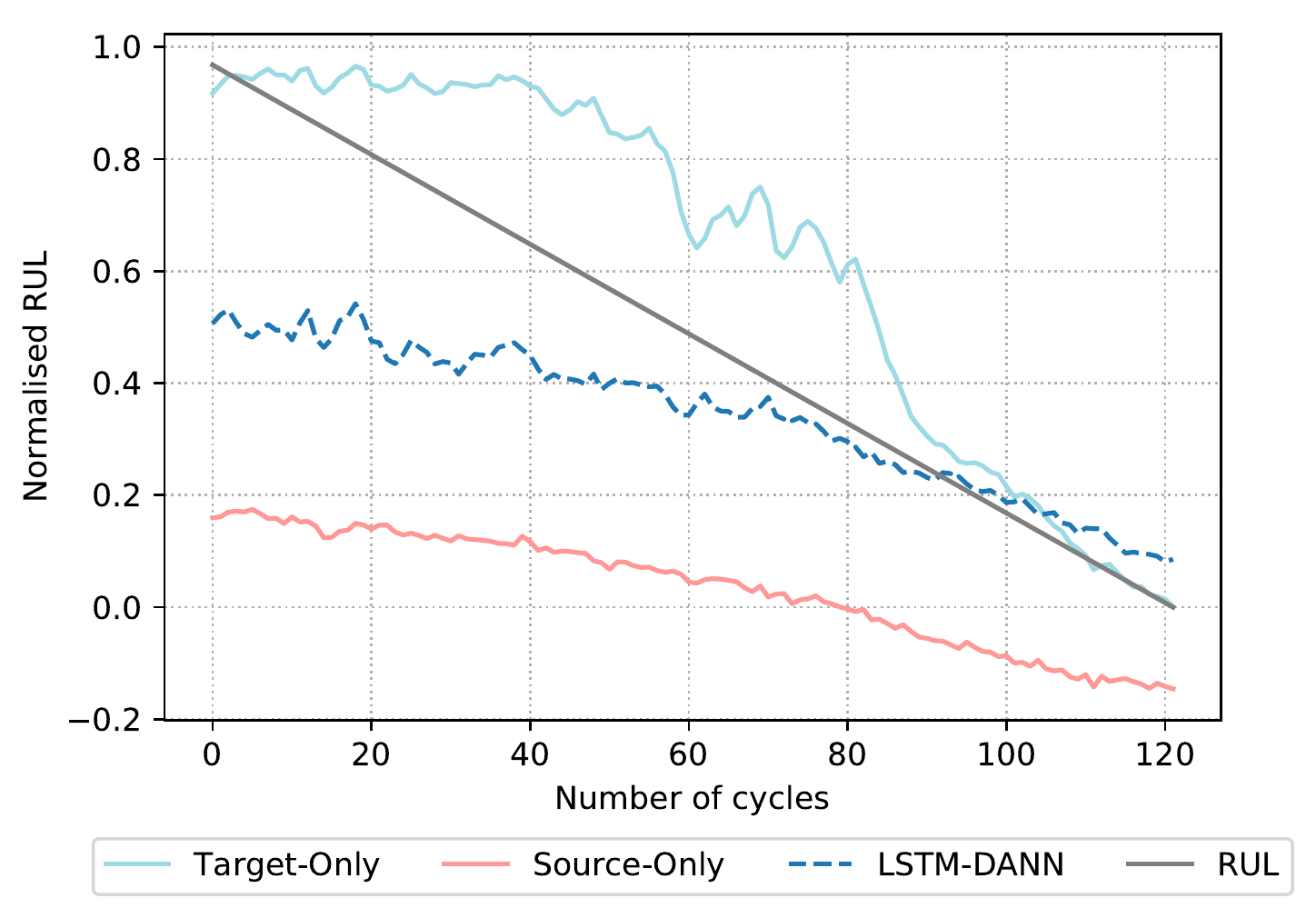}
        \caption{Target: FD003}
         \label{fig:fd01_f03}
    \end{subfigure}%
    \begin{subfigure}[b]{0.33\textwidth}
        \includegraphics[width=0.95\textwidth]{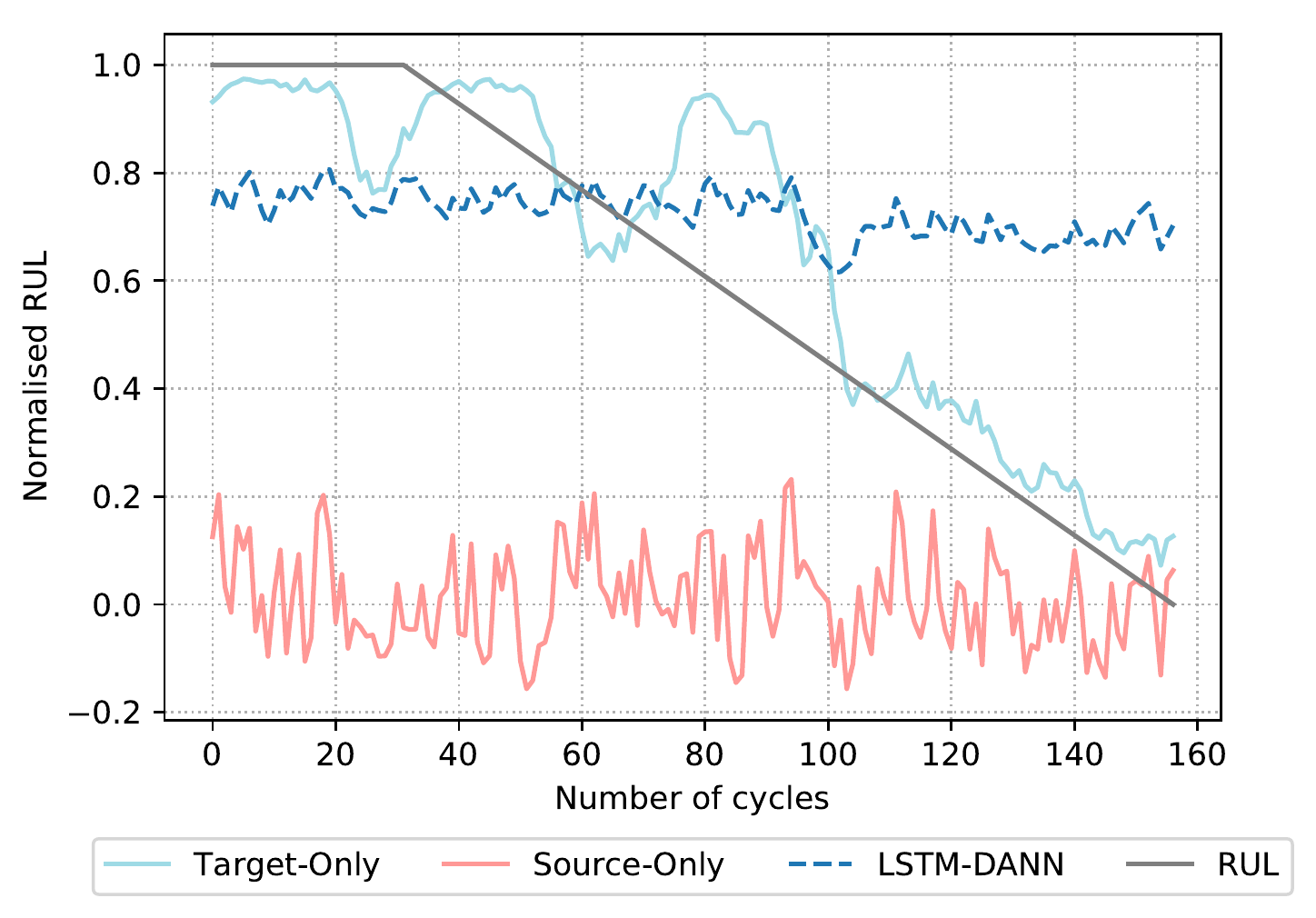}
        \caption{Target: FD004}
         \label{fig:fd01_f04}
    \end{subfigure}
    \hfill
    \caption{RUL predictions of the TARGET-ONLY, SOURCE-ONLY and LSTM-DANN models for one engine coming from the target domain cross-validation datasets.}
    \label{fig:results}
\end{figure*}

In this section, we compare the proposed model with models trained in the source domain and applied on the target domain (SOURCE-ONLY) serving as a baseline for the domain-adapted models and models trained in the target domain using target labels (TARGET-ONLY) serving as upper bound of for the proposed methods. For the models where no adaptation is performed (SOURCE-ONLY and TARGET-ONLY) we train a network of the form: ReLU(LSTM(100)) + Dropout(0.5) + ReLU(Dense(30)) + Dropout(0.1) + ReLU(Dense(20)) + Dense(1) for 100 epochs using the Adam \cite{kingma2014adam} optimiser with a learning rate of 0.001. We use an MSE loss function and $T_w$ equal to 30, 20, 30, 15 for FD001, FD002, FD003, FD004. These hyperparameters are chosen because they yield the best performances in our experiments. We present, in Table \ref{tab:comparison}, the performances in the target test datasets for each source-target pairs in our experiments. We aim to show the effect of using a model for the target domains trained only on the source domain in comparison with the proposed method. Therefore, we present the percentage change between SOURCE-ONLY and the LSTM-DANN method as $\Delta$\%. We also present the normalised RUL prediction results of several engines coming from the target cross-validation dataset in Figure \ref{fig:results}. In the figure, we present the target RUL values as well as the predictions coming from the LSTM-DANN, SOURCE-ONLY and TARGET-ONLY models. We analyse the results splitting the analysis for each domain, as its selection poses distinct difficulties in the adaptation results.
\begin{table}[h!]
\centering
\resizebox{\columnwidth}{!}{%
\begin{tabular}{cllll}
\hline
Source: FD001 & Target & SOURCE-ONLY & LSTM-DANN ($\Delta$\%) & TARGET-ONLY  \\ \hline
             - & FD002  & 71.70 $\pm$ 3.88 &  \textbf{48.62} (-32\%) $\pm$ 6.83     & 17.76 $\pm$ 0.43        \\
             - & FD003  & 51.20  $\pm$ 3.39  &  \textbf{45.87} (-10\%) $\pm$ 3.58& 12.49 $\pm$ 0.29            \\
             - & FD004  & 73.88  $\pm$ 4.50   &  \textbf{43.82} (-41\%) $\pm$ 4.15& 21.30 $\pm$ 1.06          \\ \hline
Source: FD002 & Target & SOURCE-ONLY & LSTM-DANN ($\Delta$\%) & TARGET-ONLY \\ \hline
             - & FD001  & 164.84 $\pm$ 23.00&   \textbf{28.10} (-83\%) $\pm$ 5.03& 13.64   $\pm$ 0.80       \\
             - & FD003  & 154.04 $\pm$ 21.79& \textbf{37.46 }(-76\%)  $\pm$ 1.54&   12.49   $\pm$ 0.29       \\
             - & FD004  & 37.76 $\pm$ 2.17 & \textbf{31.85} (-16\%)  $\pm$ 1.65&  21.30 $\pm$ 1.06  \\ \hline
Source: FD003 & Target & SOURCE-ONLY & LSTM-DANN ($\Delta$\%) & TARGET-ONLY \\ \hline
             - & FD001  &    49.94 $\pm$ 7.65  &    \textbf{31.74 }(-36\%) $\pm$ 9.37 & 13.64  $\pm$ 0.80        \\
             - & FD002  &    70.32 $\pm$ 4.02 &    \textbf{44.62} (-36\%) $\pm$ 1.21& 17.76 $\pm$ 0.43       \\
             - & FD004  &    69.28 $\pm$ 4.51 &    \textbf{47.94} (-31\%) $\pm$ 5.78 & 21.30 $\pm$ 1.06  \\ \hline
Source: FD004 & Target & SOURCE-ONLY & LSTM-DANN ($\Delta$\%) & TARGET-ONLY \\ \hline
             - & FD001  &    188.00 $\pm$ 25.95&   \textbf{31.54} (-83\%) $\pm$ 2.42   & 13.64  $\pm$ 0.80       \\
             - & FD002  &    \textbf{20.88} $\pm$ 1.66 &    24.93 (+19\%)  $\pm$ 1.82 &17.76  $\pm$ 0.43     \\
             - & FD003  &    157.32  $\pm$ 20.37 &    \textbf{27.84} (-82\%) $\pm$ 2.69 & 12.49  $\pm$ 0.29     \\ \hline
\end{tabular}%
}
\caption{RMSE $\pm$ Standard Deviation - Comparison between SOURCE-ONLY, TARGET-ONLY and LSTM-DANN on the test datasets.}
\label{tab:comparison}
\end{table}

\begin{enumerate}[leftmargin=*]
\item Source: FD004 \\
Several RUL prediction results for FD004 acting as source domain are presented in Figures \ref{fig:fd04_f01}, \ref{fig:fd04_f02}, \ref{fig:fd04_f03}. In the figures, one can notice that the RUL predictions of the SOURCE-ONLY model for target domains FD001 (Fig. \ref{fig:fd04_f02}) and FD003 (Fig. \ref{fig:fd04_f03}) have a large error in comparison with the observed values. On the other hand, despite some errors between the predictions and observed values, our proposed model shows higher accuracy on the target RUL values leading to a smaller error than the one from SOURCE-ONLY model. 

We also point out that for target domain FD002, the SOURCE-ONLY model already provides a good fit for the observed RUL values (Figure \ref{fig:fd04_f02}). In this case, domain adaptation results in predictions similar to SOURCE-ONLY. Usually, this result is not known beforehand, but can be expected since the marginal distribution shift across domains FD004 and FD002 is low. We also point out that FD004 is the dataset that contains 6 operating conditions and 2 fault modes. Therefore, the adaptation method can use the source domain to find correspondences between the operating conditions and fault modes in each source-target pair. That is of practical value since one could use previous data coming from different conditions and fault modes to estimate better RUL values on unobserved run-to-failure data.

\item Source: FD003 \\
In the examples provided, the adaptation from FD003 to FD002 (Fig. \ref{fig:fd03_f02}) and FD004 (Fig. \ref{fig:fd03_f04}) show worse results than the one for FD001 (Fig. \ref{fig:fd03_f01}). This is the case as FD003 is much more similar to FD001, varying only the number of fault modes. As it can be seen on Table \ref{tab:comparison} the LSTM-DANN model yields a lower RMSE value when compared to the SOURCE-ONLY model showing that the weights learned by the network can be effectively used even when the domains are already similar.

For target domains FD002 and FD004, the differences between domains are more prominent. FD003 has one operating condition while FD002 and FD004 have 6 different operating conditions and different sensor values near a failure. Despite the difficulties in transferring from such distinct domains, the domain adaptation method can improve the SOURCE-ONLY model showing a better prediction error in the test dataset. However, some higher errors are present as the estimated values are noisy and do not fit the linear degradation model in its complete extension.

\item Source: FD002 \\
In Figures \ref{fig:fd02_f01}, \ref{fig:fd02_f03} and \ref{fig:fd02_f04} we present engines from the FD001, FD003, FD004 cross-validation target datasets. Similar to the inverted experiment, the similarities between the FD004 and FD002 make the SOURCE-ONLY model fit the target data with high accuracy. In this experiment pair, our model is also able to fit the target data with a similar error level as the one without adaptation. On the other hand, we can improve the predictions on the FD001 target dataset in comparison to SOURCE-ONLY. In this case, FD001 and FD002 share the same fault mode (HPC degradation), but FD002 has more operating conditions than FD001 which makes the algorithm able to learn the degradation function better than a model without adaptation. Our method can also produce more accurate results than SOURCE-ONLY when the target domain is FD003. Although both operating conditions and fault mode are different across domains, the predictions produce lower errors and a better fit to the linear degradation model. The result of this experiment shows that it is possible to transfer from a domain that has more operating conditions than the target domains under the same or different fault modes. This is important, as in practice one is interested in reusing previous gathered or simulation data to predict the RUL on unseen data.

\item Source: FD001 \\
In our experiments, FD001 presents the highest errors when functioning as a source domain. When target domains are FD002 of FD004 (Figs. \ref{fig:fd01_f02} and \ref{fig:fd01_f02}) the best solution found is one that yields a flattened curve over the entire cycle. It can be observed that the SOURCE-ONLY model has a similar behaviour when being used to predict the target dataset. However, the proposed model is capable of adjusting the learned values towards a mean RUL value. Also, we point out that FD001 is the dataset containing only one operating condition and fault mode. That is, we are trying to transfer to domains where the fault modes and conditions are not in the source domain. Similar to other results, this shows to be a much harder problem to the methodology proposed.  When the target domain is FD003 (Fig. \ref{fig:fd01_f03}), we are attempting to learn in a domain with one fault mode and predict in a domain with two fault modes and similar operating conditions. For this case, the model results in RUL prediction curves that can fit the trend of the observed RUL values to a lower RMSE than the SOURCE-ONLY model. 
\end{enumerate}

We notice, in Table \ref{tab:comparison}, that the proposed methodology is capable of improving performance over almost all but one SOURCE-ONLY methods in our experiments. The results of the adapted models change depending on the information contained in the dataset acting as source domain. We achieve better results when FD004 dataset acts as source domain as it contains all 6 operating conditions and 2 fault modes. Also, when the distribution shift in the source and target domains are similar, a model trained in the source domain with no adaptation can already achieve strong performance in the learning task. 

In particular, using LSTM-DANN does not add much value when the distributions between domains are very similar and the source domain contains more operating conditions and fault modes than the target domain (e.g. FD004 to FD002). For this example, even if we have access to the ground truth values in the target domain the performance (RMSE) among SOURCE-ONLY, LSTM-DANN and TARGET-ONLY models will not be considerably improved, unlike other experiments. Moreover, among our 10 runs there were cases when the LSTM-DANN would be able to find better RMSE values than the ones found by SOURCE-ONLY indicating that there may be better hyperparameters than the ones proposed in this paper that would result in lower RMSE values.  

For the other cases, the results show that when the source data ``contain" the target data the adaptation achieve lower RMSE results and a better fit. It is expected that the observed features in the source domain can be used to improve predictions in the target domain; thus, having a source domain with similar degradation data helps to learn. On the other hand, learning from a domain with fewer operating conditions and fault modes (e.g. FD001) to one with more conditions and fault modes proves to be a harder task. The model can adjust for the mean RUL adaptation value but fails to learn the linear degradation model that emerges from the target domain. However, the results still prove to be better estimations of the RUL in a target domain when compared to SOURCE-ONLY models.

As a rule of thumb, the results presented show that the methodology could be useful whenever one has similar degradation data with a degree of distribution shift across domains. The degree to which the distributions vary can determine how accurate an adaptation could result, but further investigation has to be done to provide a reliable estimate of ``when" to transfer. The model achieves the best results when enough source data under different conditions has been observed. In practice, one could use simulated data acting as a source domain with various fault scenarios. In many applications, real-world data is dissimilar to simulation data due to noise and unexpected sensor behaviour. To improve the models, one could use the LSTM-DANN method to adapt from simulated data to real-world data and achieve better results in RUL predictions.

\subsection{Comparison to Domain Adaptation Approaches}

To assess the quality of the model in transferring the degradation patterns from a source to a target domain, we test several well-known methodologies for unsupervised domain adaptation. Two different methodologies are carried out, Transfer Component Analysis (TCA) and CORrelation ALignment (CORAL).
\begin{enumerate}
    \item TCA \\
Transfer Component Analysis \cite{pan2011domain} is a well-known unsupervised domain adaptation method that focuses on finding a common shared feature representation between source and target domains. It transfers components across domains in a Reproducing Kernel Hilbert Space (RKHS) using Maximum Mean Discrepancy (MMD) and different kernels to construct a feature space that minimises the difference between the domains. We use the TCA feature representation to train a shallow neural network with one fully connected layer and 32 units (TCA-NN) and a deep feed forward neural network (TCA-DNN) containing the same amount of layers and units as our final models. In our tests, we apply the radial basis function (RBF) kernels extracting 20 transfer components with $\lambda = 1$ and $\gamma = 1$.

\item{CORAL} \\
CORrelation ALignment (CORAL) \cite{sun2016return} is a metric that minimises domain shift by aligning the second-order statistics of a source and target distributions, without requiring target labels. After the alignment is found the new feature space can be used to train a model in the transformed source domain. Similar to the TCA method we use such a method in combination with a shallow (CORAL-NN) and deep neural network architectures (CORAL-DNN).
\end{enumerate}

For all compared methods in this study, the input data are the same, the mean squared error (MSE) is used as the loss function and the Adam \cite{kingma2014adam} optimisation algorithm is applied with a learning rate of 0.001.

TCA and CORAL provide out-of-the-box methods that can be easily applied in a domain adaptation setup where no observed values are available.  Both methods are not entirely suitable for temporal data,  for this reason we perform domain adaptation on TCA and CORAL using features coming from time $t-1$ and RUL at time $t$.  After the new features are computed we train a FFNN model using the same number of units as in the SOURCE-ONLY models.  The results are summarised for the target cross-validation datasets (not seen during training)  in Table \ref{rmse-table}. It can be seen that on average our proposed method yields lower RMSE than the methodologies tested in all but one experiment pair. This supports that the proposed deep adaptation architecture is well suited for the studied prognostics prediction problem. 

The LSTM structure in combination with the adversarial classification loss are capable of extracting useful temporal features from multivariate time series to perform adaptation from source and target domain pairs. In other terms, this means that the methodology performs better than the tested \textit{out-of-the-box }adaptation methods not tailored for time-series sensor data. The results provide a foundation for using the proposed domain adaptation method in cases when one has limited observed RUL data in one domain, but is concerned with predicting RUL targets in a similar domain with distinct operating conditions and fault modes.
\begin{table}[h!]
\centering
\resizebox{\columnwidth}{!}{%
\begin{tabular}{clllllll}
\hline
Source: FD001 & Target & TCA-NN & TCA-DNN & CORAL-NN & CORAL-DNN & LSTM-DANN \\ \hline
             - & FD002  &  94.1 $\pm$ 1.0 &  90.0 $\pm$ 2.9 &99.2 $\pm$ 3.6&77.5 $\pm$ 4.6&    \textbf{46.4} $\pm$ 3.6 \\
             - & FD003  & 120.0 $\pm$ 1.0 & 116.1 $\pm$ 1.0 &60.0 $\pm$ 0.7&69.6 $\pm$ 5.2&   \textbf{37.3} $\pm$ 3.4  \\
             - & FD004  & 120.1 $\pm$ 1.0 & 113.8 $\pm$ 6.9 &107.7 $\pm$ 2.8&84.6 $\pm$ 7.0&  \textbf{43.5} $\pm$ 5.3       \\ \hline
Source: FD002 & Target & TCA-NN & TCA-DNN & CORAL-NN & CORAL-DNN & LSTM-DANN \\ \hline
             - & FD001  &  94.7  $\pm$ 1.1 &85.6 $\pm$ 5.5&77.9 $\pm$ 19&80.9 $\pm$ 9.4&   \textbf{31.2} $\pm$ 5.4\\
             - & FD003  & 107.4 $\pm$ 3.7&111.5 $\pm$ 7.2&60.9 $\pm$ 15.1&79.8 $\pm$ 10.1&   \textbf{32.2} $\pm$ 3.1\\
             - & FD004  &  93.5 $\pm$ 2.8 & 94.4 $\pm$ 6.7&37.5 $\pm$ 0.5&43.6 $\pm$ 3.6&   \textbf{27.7} $\pm$ 1.5\\ \hline
Source: FD003 & Target & TCA-NN & TCA-DNN & CORAL-NN & CORAL-DNN & LSTM-DANN \\ \hline
             - & FD001  &  98.7  $\pm$ 0.4 & 90.5  $\pm$ 4.6 &26.5 $\pm$ 0.5 &\textbf{26.5} $\pm$ 1.9&  30.6 $\pm$ 6.2         \\
             - & FD002  &  90.5  $\pm$ 0.3 & 80.8  $\pm$ 4.3 &113.2 $\pm$ 4.5&75.6 $\pm$ 9.5&   \textbf{43.1} $\pm$ 1.4       \\
             - & FD004  &  78.9  $\pm$ 5.3 & 102.6 $\pm$ 3.2 &113.9 $\pm$ 5.5&77.2 $\pm$ 9.1&  \textbf{49.7} $\pm$ 9.1  \\ \hline
Source: FD004 & Target & TCA-NN & TCA-DNN & CORAL-NN & CORAL-DNN  & LSTM-DANN \\ \hline
             - & FD001  &  98.5 $\pm$ 0.4 &  85.6 $\pm$ 5.0 & 119.1 $\pm$ 16.7 & 94.0 $\pm$ 8.8   & \textbf{25.4} $\pm$ 4.2\\  
             - & FD002  &  75.3 $\pm$ 1.7 &  80.8 $\pm$ 5.8 &  37.3 $\pm$ 0.6  & 30.9 $\pm$ 1.4   & \textbf{26.9} $\pm$ 3.3\\
             - & FD003  &  77.2 $\pm$ 6.0 & 102.9 $\pm$ 2.7 &  68.1 $\pm$ 11.1 & 68.6 $\pm$ 11.2  & \textbf{23.6} $\pm$ 5.0 \\ \midrule   
\end{tabular}%
}
\caption{RMSE $\pm$ Standard Deviation - Domain Adaptation Models on target cross-validation data.}
\label{rmse-table}
\end{table}
\subsection{Comparison to Non-adapted RUL Prediction Approaches Using  Target Domain Labels}
We provide the results of TARGET-ONLY models in Table \ref{tab:comparison} as a the best case performance where the RUL target values are used to train the models. That is, these are the best results found if we could use the RUL values in the target domain for training. We also present a comparison of the TARGET-ONLY models to the state-of-the-art results in the C-MAPPS datasets for the RMSE and Scoring performance metrics in Tables \ref{tab:comparison-lit-rmse} and \ref{tab:comparison-lit-score}. We compare our method to the ones in the literature to attest that our model can show similar results when presented with a complete labelled dataset as others methods proposed in the literature. 

\begin{table}[h!]
\centering
\resizebox{\columnwidth}{!}{
    \begin{tabular}{clllll}
    \hline
    Dataset & TARGET-ONLY & GA + LSTM \cite{ListouEllefsen2019} & CNN + FFNN \cite{Li2018} & MODBNE \cite{zhang2017multiobjective} &LSTM + FFNN  \cite{Zheng2017}\\ \hline
                 FD001 &  13.64 & \textbf{12.56}&  12.61& 15.04 & 16.14      \\
                 FD002 & \textbf{17.76} & 22.73&  22.36 & 25.05 & 24.49    \\
                 FD003 & 12.49  & \textbf{12.10}&  12.64 & 12.51& 16.18   \\
                 FD004 & \textbf{21.30} & 22.66 &  23.31 & 28.66&  28.17  \\\hline
    \end{tabular}%
}
\caption{RMSE comparison between TARGET-ONLY and other models on the literature on the C-MAPPS datasets}
\label{tab:comparison-lit-rmse}
\end{table}

We compare our TARGET-ONLY model with the LSTM methods proposed by \citet{ListouEllefsen2019} (GA + LSTM) and \citet{Zheng2017} as (LSTM + FFNN). We also compare to the Convolutional Neural Network (CNN + FFNN) methodology proposed by \citet{Li2018} and the Multiobjective Deep Belief Networks Ensemble (MODBNE) proposed by \citet{zhang2017multiobjective}. In Table \ref{tab:comparison-lit-rmse} we report the results of our experiments and the results reported in the original papers of the compared approaches. We notice that our models provides the best known RMSE results for datasets FD002 and FD004 showing that the method proposed can yield a strong performance on the datasets with more operating conditions. For datasets FD001 and FD002 we can produce similar results to the best known performance models proposed in the literature for both RMSE and Scoring metrics. We note that these methods cannot be directly used for dataset without labels as they are trained and tested in the same domain.
\begin{table}[h!]
\centering
\resizebox{\columnwidth}{!}{%
\begin{tabular}{clllll}
\hline
Dataset & TARGET-ONLY & GA + LSTM \cite{ListouEllefsen2019} & CNN + FFNN \cite{Li2018} & MODBNE \cite{zhang2017multiobjective} &LSTM + FFNN \cite{Zheng2017} \\ \hline
             FD001 &  300 & \textbf{231}&  274& 334 & 338      \\
             FD002 & \textbf{1,638}  & 3,366&  10,412 & 5,585 & 4,450    \\
             FD003 & 267  & \textbf{251}&  284 & 422& 852   \\
             FD004 & 2,904 & \textbf{2,840} &  12,466 & 6,558&  5,550  \\\hline
\end{tabular}%
}
\caption{Scoring comparison between TARGET-ONLY and other models on the literature on the C-MAPPS datasets}
\label{tab:comparison-lit-score}
\end{table}

\section{Discussion}

\subsection{Relationship to Standardisation}

A common straightforward way to attempt to align a source and target domains and reduce the difference between the means and variances of the input distributions is performing standardisation. That is, local mean centering (subtracting the mean) and dividing by the standard deviation of each input feature. This leads to each feature in the data to have zero-mean (moment matching) and unit-variance. 

To test whether such transformation already suffices as an alignment strategy we standardise the data before feeding it to SOURCE-ONLY and TARGET-ONLY architectures. We select FD004 as source domain as results have shown that adaptation is possible for all remaining C-MAPPS datasets. We present the RMSE values for each methodology evaluated in the test target datasets in Table \ref{standard-table}. We compare to the original LSTM-DANN trained on the data with the min-max normalisation and to a model trained on the standardised data (LSTM-DANN-STD) and to reference models SOURCE-ONLY and TARGET-ONLY.

\begin{table}[h!]
\centering
\resizebox{\columnwidth}{!}{%
\begin{tabular}{cllllll}
\hline
Source: FD004 & Target & SOURCE-ONLY-STD & LSTM-DANN & LSTM-DANN-STD & TARGET-ONLY-STD   \\ \hline
             - & FD001  &  53.31 $\pm$ 5.02 &  \textbf{31.54} $\pm$ 2.42 & 32.62$\pm$ 2.07 & 14.51 $\pm$ 1.55   \\  
             - & FD002  &  23.22 $\pm$ 1.01 &  24.93  $\pm$ 1.82  &  \textbf{21.78} $\pm$ 1.71  & 18.44 $\pm$ 0.42   \\
             - & FD003  &  62.76 $\pm$ 7.52 &  \textbf{27.84} $\pm$ 2.69 &  40.20 $\pm$ 7.03 & 16.03 $\pm$ 0.33  \\ \midrule   
\end{tabular}%
}
\caption{Test performance (RMSE $\pm$ Standard Deviation) of SOURCE-ONLY, TARGET-ONLY, LSTM-DANN-STD. Models are trained on standardised training data with zero-mean and unit-variance.}
\label{standard-table}
\end{table}

\begin{figure*}[h!]
    \setlength{\belowcaptionskip}{-12pt}
    \centering
    
    Source: FD004
    \begin{subfigure}[b]{0.33\textwidth}
        \includegraphics[width=\textwidth]{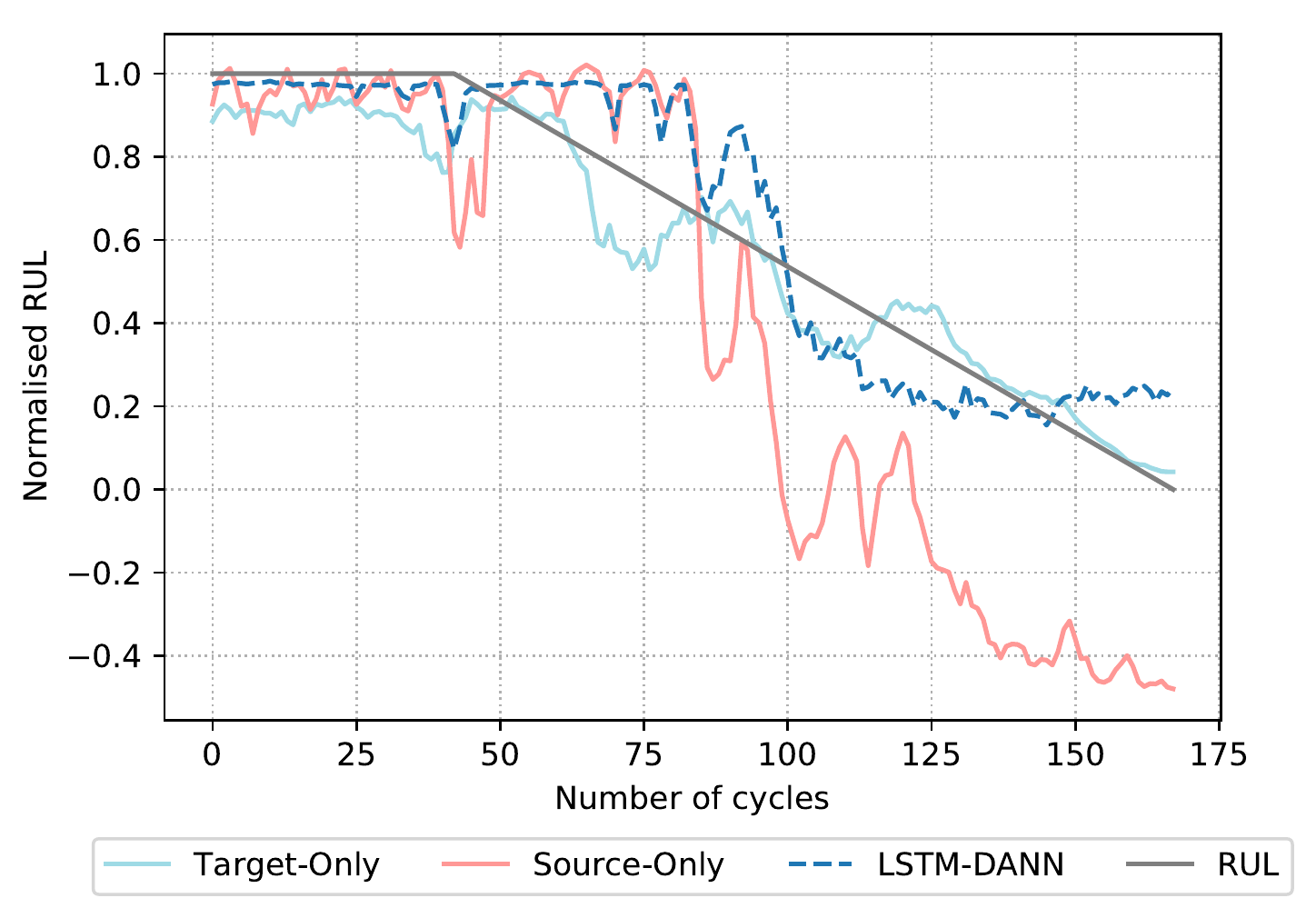}
        \caption{Target: FD001}
         \label{fig:fd04_f01_std}
    \end{subfigure}%
    \begin{subfigure}[b]{0.33\textwidth}
        \includegraphics[width=\textwidth]{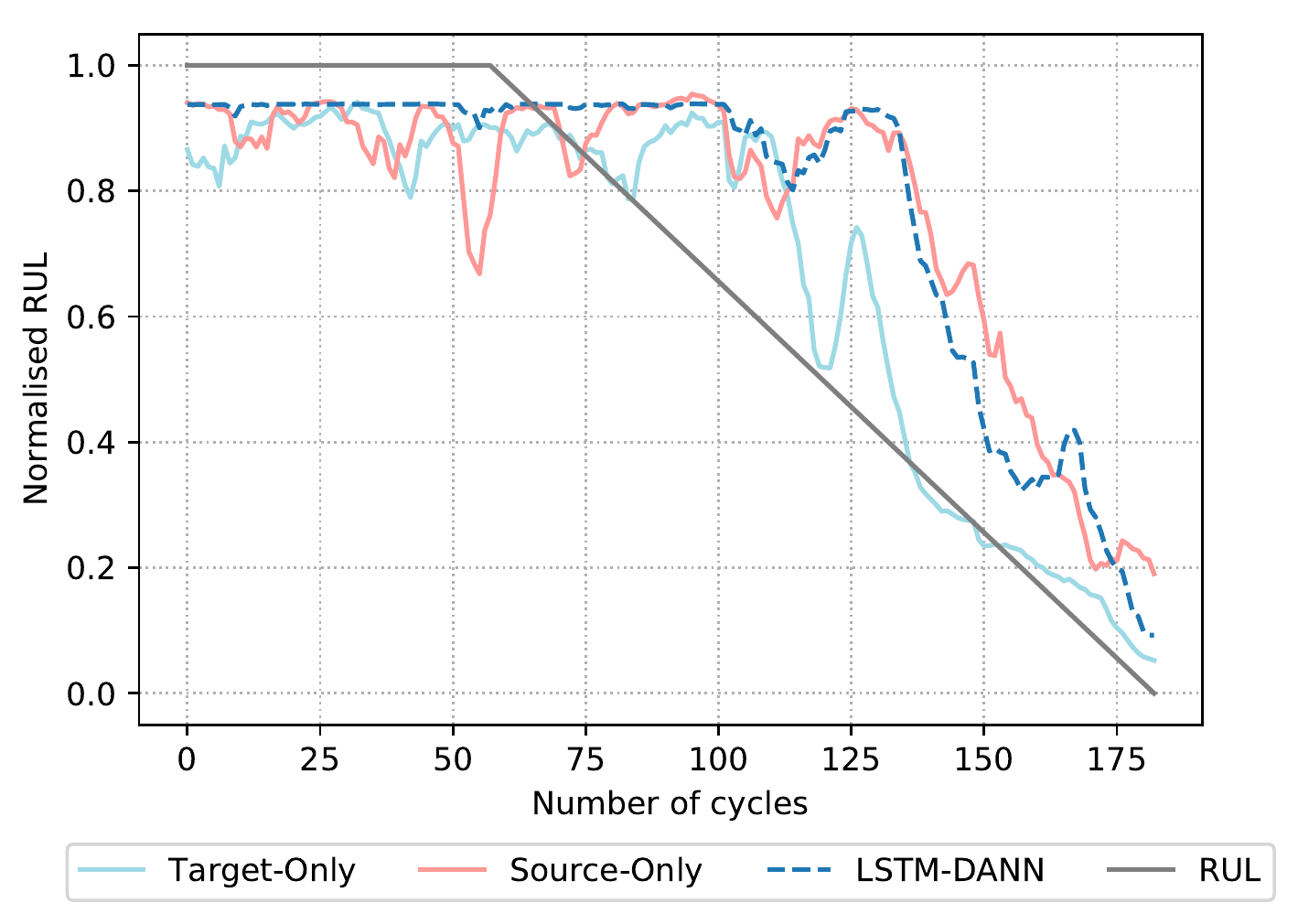}
        \caption{Target: FD002}
         \label{fig:fd04_f02_std}
    \end{subfigure}%
    \begin{subfigure}[b]{0.33\textwidth}
        \includegraphics[width=\textwidth]{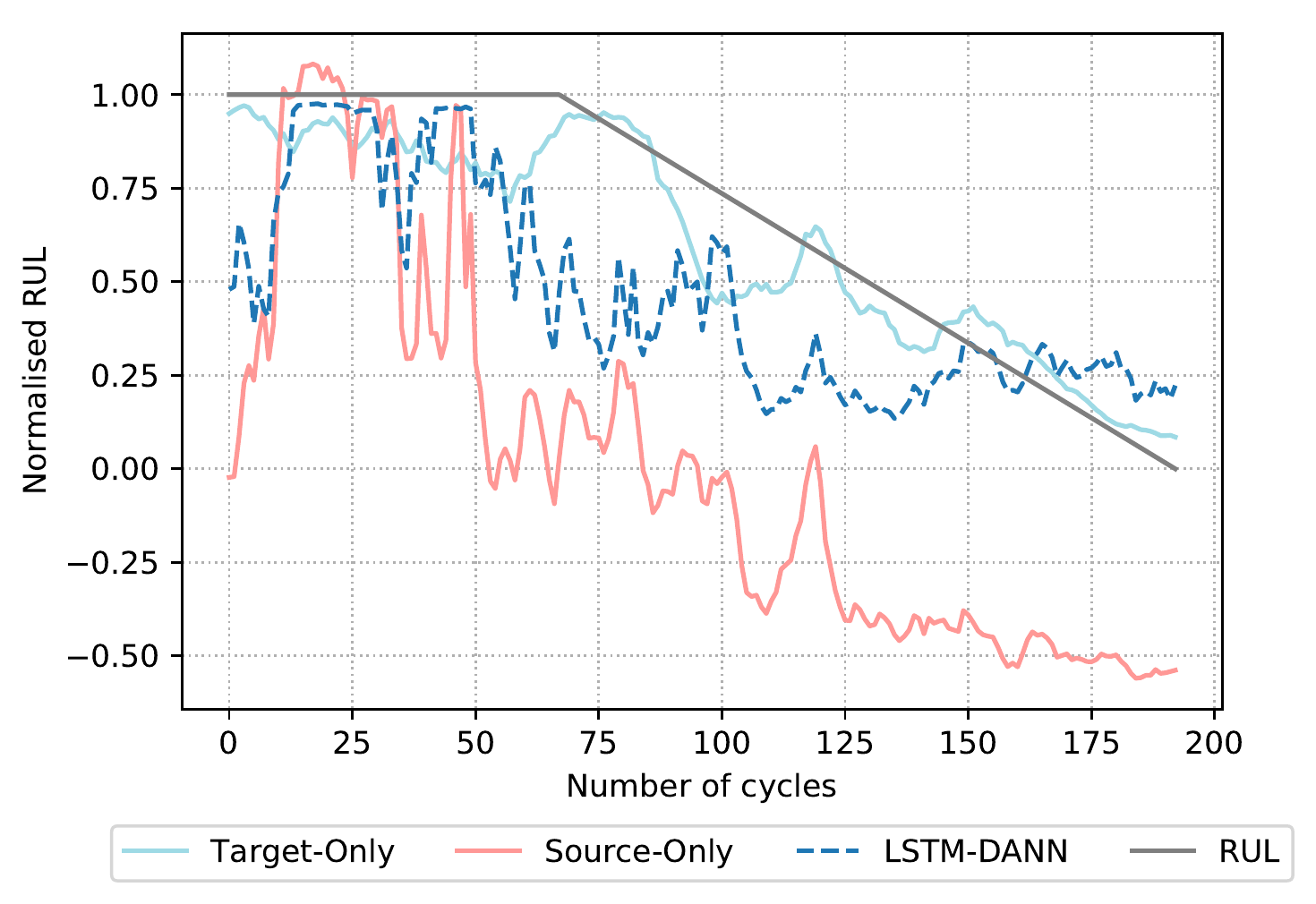}
        \caption{Target: FD003}
         \label{fig:fd04_f03_std}
    \end{subfigure}
 
    \hfill
    \caption{RUL predictions of the TARGET-ONLY, SOURCE-ONLY and LSTM-DANN-STD models in the standardised target cross-validation datasets.}
    \label{fig:std_results}
\end{figure*}

The results in Table \ref{standard-table} show that on average the RMSE performances of SOURCE-ONLY models are considerably improved when the data is normalised with zero mean and unit-variance. However, our proposed methodology (LSTM-DANN-STD) still outperforms the baseline models and provide a better fit to the target data (Figure \ref{fig:std_results}). We point out that training on standardised data causes the LSTM-DANN-STD models to saturate and start overfitting much faster than on previous experiments. This effect negatively impacted the adaptation performance on FD003 as training progressed for more epochs than necessary. To address this effect, simple changes in the models hyperparameters could be made. For example one could consider a lowering the value of $\alpha$  and reducing learning rates. 

\section{Conclusion}
In this paper, a deep learning method for domain adaptation in prognostics is proposed based on a Long-Short Term Memory Neural Network and a Domain Adversarial Neural Network (LSTM-DANN). We use time windows to incorporate long term sequences in the feature extraction layers. Experiments are carried out on the popular C-MAPPS dataset to show the effectiveness of the proposed method. The goal of the task is to estimate the remaining useful lifetime for aircraft engines units accurately while transferring from a source domain with observed RUL values to a target domain with only input features.

We normalise the datasets independently to allow for higher distribution shifts across domains. It is worth noticing that locally normalising the data to zero-mean and unit variance improves the performance of SOURCE-ONLY methods on the target datasets. Nevertheless, we focus on using the raw features and an adequate training procedure to find the weights representation that can accommodate the original distribution shift between domains. In general, we are capable to achieve lower errors between the prediction and the actual RUL value in comparison to a model with no adaptation features. We notice that the RUL prediction can be more effectively transferred from datasets that have more fault modes or operating conditions than their target counterpart. On the other hand, transferring from a dataset with fewer operating conditions and fault modes to a higher number of characteristics is a much harder task. However, even in the harder latter scenario the model is able to correct the RUL predictions to alleviate the RMSE error without utilising the target RUL values. Furthermore, the prognostic results obtained by the proposed method are compared with the out-of-the-box domain adaptation models. In our tests, the proposed network has shown superior performance in comparison to other simple domain adaptation methods. We point out that no thorough evaluation of domain adaptation methods was performed as most of the domain adaptation methodologies do not focus on sequential temporal data. Instead, we focused on methods that would require little domain adaptation knowledge and architecture tweaking for comparison. 

Additionally, it should be noted that in common real-world online applications, the whole life-span data are usually not available. As future research, we argue that domain adaptation methods should be able to accommodate non-complete data coming from the target domain. As it is the case in many real-world scenarios, the sooner an accurate prediction can be done on the RUL of an equipment means that early actions can be planned to prevent equipment downtime. Furthermore, the network could be made such that it can retain hidden states for a longer period of time, by allowing states flow between mini-batches of data. Such modification would have allowed the network to "remember" for longer periods of time while training. This modification comes with a cost, padding and number of training examples within a batch have to be carefully selected to allow states to propagate between batches. In our experiments the varying sizes of the time sequences and number of training examples between domains led to mini-batches of a small size, which made training particularly slow.

While promising experimental results have been obtained by the proposed method, further architecture optimisation is still necessary, as hyperparameter search is restrained. Also, deep learning methods generally suffer from high computing burden, thus using larger datasets could be made computationally intractable. In addition, we report that the Scoring function is only used for evaluation purposes. No optimisation steps are taken towards the minimisation of this function. Further development could incorporate the function in a learning algorithm.

\section{Acknowledgments}
This work was supported by the ``Netherlands Organisation for Scientific Research" (NWO). Project: NWO Big data - Real Time ICT for Logistics. Number: 628.009.012

\section{References}

\bibliographystyle{model1-num-names}
\bibliography{sample.bib}







\end{document}